\documentclass{article} % For LaTeX2e
\usepackage{iclr2018_conference,times}

\usepackage{natbib}

\usepackage[utf8]{inputenc} % allow utf-8 input
\usepackage[T1]{fontenc}    % use 8-bit T1 fonts
\usepackage{url}            % simple URL typesetting
\usepackage{booktabs}       % professional-quality tables
\usepackage{amsmath}
\usepackage{amssymb}
\usepackage{amsfonts}       % blackboard math symbols
\usepackage{nicefrac}       % compact symbols for 1/2, etc.
\usepackage{microtype}      % microtypography
\usepackage{subcaption}
\usepackage[pdftex]{graphicx}
 \usepackage{booktabs} % Pretty tables.
\usepackage{comment}
\usepackage{xspace}
\usepackage{color} % Colored text.
\usepackage[font={footnotesize},labelfont={footnotesize}]{caption} % Small sized  captions.
\usepackage{wrapfig}

\usepackage[pdfauthor={Kevin Murphy},%
colorlinks=true,%
pdftex]{hyperref}

\usepackage[noabbrev,capitalize,nameinlink]{cleveref}

\usepackage{multirow}
\usepackage{color, colortbl}
\definecolor{Gray}{gray}{0.95}

\newcommand{\myvec}[1]{\mathbf{#1}}
\newcommand{\myvecsym}[1]{\boldsymbol{#1}}

\newcommand{\vzero}{\myvecsym{0}}

\newcommand{\vmu}{\myvecsym{\mu}}

\newcommand{\vphi}{\myvecsym{\phi}}

\newcommand{\vtheta}{\myvecsym{\theta}}

\newcommand{\vx}{\myvec{x}}

\newcommand{\vy}{\myvec{y}}

\newcommand{\vz}{\myvec{z}}
\newcommand{\vC}{\myvec{C}}

\newcommand{\vI}{\myvec{I}}

\newcommand{\vX}{\myvec{X}}
\newcommand{\mymathcal}[1]{\mathcal{#1}}
\newcommand{\calL}{\mymathcal{L}}
\newcommand{\calM}{\mymathcal{M}}

\newcommand{\calO}{\mymathcal{O}}

\newcommand{\calS}{\mymathcal{S}}

\newcommand{\calX}{\mymathcal{X}}
\newcommand{\calY}{\mymathcal{Y}}

\newcommand{\expect}[1]{\mathbb{E}\left[{#1}\right]}
\newcommand{\expectQ}[2]{\mathbb{E}_{{#2}}\left[ {#1} \right]}

\newcommand{\entropy}{\mathbb{H}}

\newcommand{\KL}{\mathrm{KL}}
\newcommand{\JS}{\mathrm{JS}}

\newcommand{\yobs}{\vy_{\calO}}
\newcommand{\ymiss}{\vy_{\calM}}

\newcommand{\ie}{\emph{i.e.}}
\newcommand{\eg}{\emph{e.g.}}
\newcommand{\cf}{\emph{c.f.}}
\newcommand{\vs}{\emph{vs.}}

\newcommand{\eat}[1]{}
\newcommand{\remark}[2]{} %{\textcolor{blue}{#1}: \textcolor{blue}{#2}}
\newcommand{\replace}[2]{} %{\sout{#1} \textbf{#2}}

\newcommand{\todo}[1]{\marginpar{\tiny{\textcolor{red}{#1}}}}
\newcommand{\TODO}{\todo}

\newcommand{\rama}{\textcolor{black}}
\newcommand{\ramacr}{}

\newcommand{\jon}[1]{} %{\remark{Jon}{#1}}
\newcommand{\elbo}{\mathrm{elbo}}

\newcommand{\data}{\mathcal{D}}

\newcommand{\gauss}{\mathcal{N}}

\newcommand{\correctness}{correctness\xspace}

\newcommand{\coverage}{coverage\xspace}
\newcommand{\ind}[1]{\mathbb{I}(#1)}

\newcommand{\obs}{{\mathcal O}}
\newcommand{\missing}{{\mathcal M}}
\newcommand{\attributes}{{\mathcal A}}
\newcommand{\mask}[2]{#1_{#2}}
\newcommand{\concept}{\mask{\vy}{\obs}}
\newcommand{\extension}{{\mathcal S}}

\newcommand{\pdata}{\hat{p}}
\newcommand{\bivcca}{BiVCCA\xspace}

\newcommand{\jmvae}{JMVAE\xspace}
\newcommand{\JMVAE}{\jmvae}
\newcommand{\telbo}{TELBO\xspace}
\newcommand{\CelebA}{CelebA\xspace}
\newcommand{\CELEBA}{CelebA\xspace}
\newcommand{\MNISTa}{MNIST-A\xspace}
\newcommand{\MNISTA}{\MNISTa}
\newcommand{\mnistaffine}{\MNISTa}
\newcommand{\mnistbit}{MNIST-2bit\xspace}
\newcommand{\iid}{iid\xspace}
\newcommand{\comp}{comp\xspace}

\newcommand{\unimodalx}{\lambda_x^x}
\newcommand{\bimodalx}{\lambda_x^{xy}}
\newcommand{\unimodaly}{\lambda_y^y}
\newcommand{\bimodaly}{\lambda_y^{yx}}

\newcommand{\pp}{p_{\vtheta}}
\newcommand{\px}{p_{\vtheta_x}}
\newcommand{\py}{p_{\vtheta_y}}
\newcommand{\qq}{q_{\vphi}}
\newcommand{\qx}{q_{\vphi_x}}
\newcommand{\qy}{q_{\vphi_y}}
\newcommand{\qqavg}{q_{\vphi}^{\mathrm{avg}}}

\title{Generative Models of Visually Grounded Imagination}

\newcommand*{\myfont}{\fontfamily{cmtt}\selectfont}

\author{
Ramakrishna Vedantam\thanks{Work performed during an internship at Google.}\\
Georgia Tech\\
{\myfont vrama@gatech.edu}\\
\And
Ian Fischer\\
Google Inc.\\
{\myfont iansf@google.com}\\
\And
Jonathan Huang\\
Google Inc.\\
{\myfont jonathanhuang@google.com}\\
\And
Kevin Murphy\\
Google Inc.\\
{\myfont kpmurphy@google.com}\\
}

\iclrfinalcopy % Uncomment for camera-ready version, but NOT for submission.

\begin{document}

\maketitle

\begin{abstract}
  It is easy for people  to imagine what a man with pink hair looks like,
  even if they have never seen such a person before.
We call the ability to create images of novel semantic concepts
\emph{visually grounded imagination}.
In this paper, we show how we can modify
variational auto-encoders to perform this task.
Our method uses a novel training objective,
and a novel  product-of-experts inference network,
which can  handle partially specified (abstract) concepts in a
principled and efficient way.
We also propose a set of easy-to-compute evaluation metrics
that capture our intuitive notions of what it means to have good
visual imagination, namely correctness, coverage, and
compositionality (the \emph{3 C’s}).
Finally, we perform  a detailed comparison
of our method with two existing joint image-attribute VAE methods
(the \jmvae  method of \citet{Suzuki2017}
and the \bivcca method of \citet{Wang2016}) by applying them to
 two  datasets:  the MNIST-with-attributes dataset (which we introduce here),
 and the \CelebA dataset~\citep{CelebA}.
\end{abstract}

\section{Introduction}
\label{sec:intro}

Consider the following two-party communication game:
a speaker thinks of a visual concept $C$, such as
``men with black hair'',
and then generates a description $\vy$ of this concept,
which she sends to a listener;
the listener interprets the description $\vy$,
by  creating an internal representation $\vz$,
which captures its ``meaning''.
We can think of $\vz$ as representing
a set of ``mental images'' which depict the concept $C$.
To test whether the listener has correctly ``understood'' the concept,
we ask him to draw a set of real images $\calS = \{ \vx_s: s=1:S\}$,
which depict the concept $C$.
He then sends these back to the speaker,
who checks to see if the images correctly
match the concept $C$.
We call this process
%of generating images of visual concepts
{\em visually grounded imagination}.

\begin{figure}[ht]
	\centering
	\includegraphics[width=0.6\linewidth]{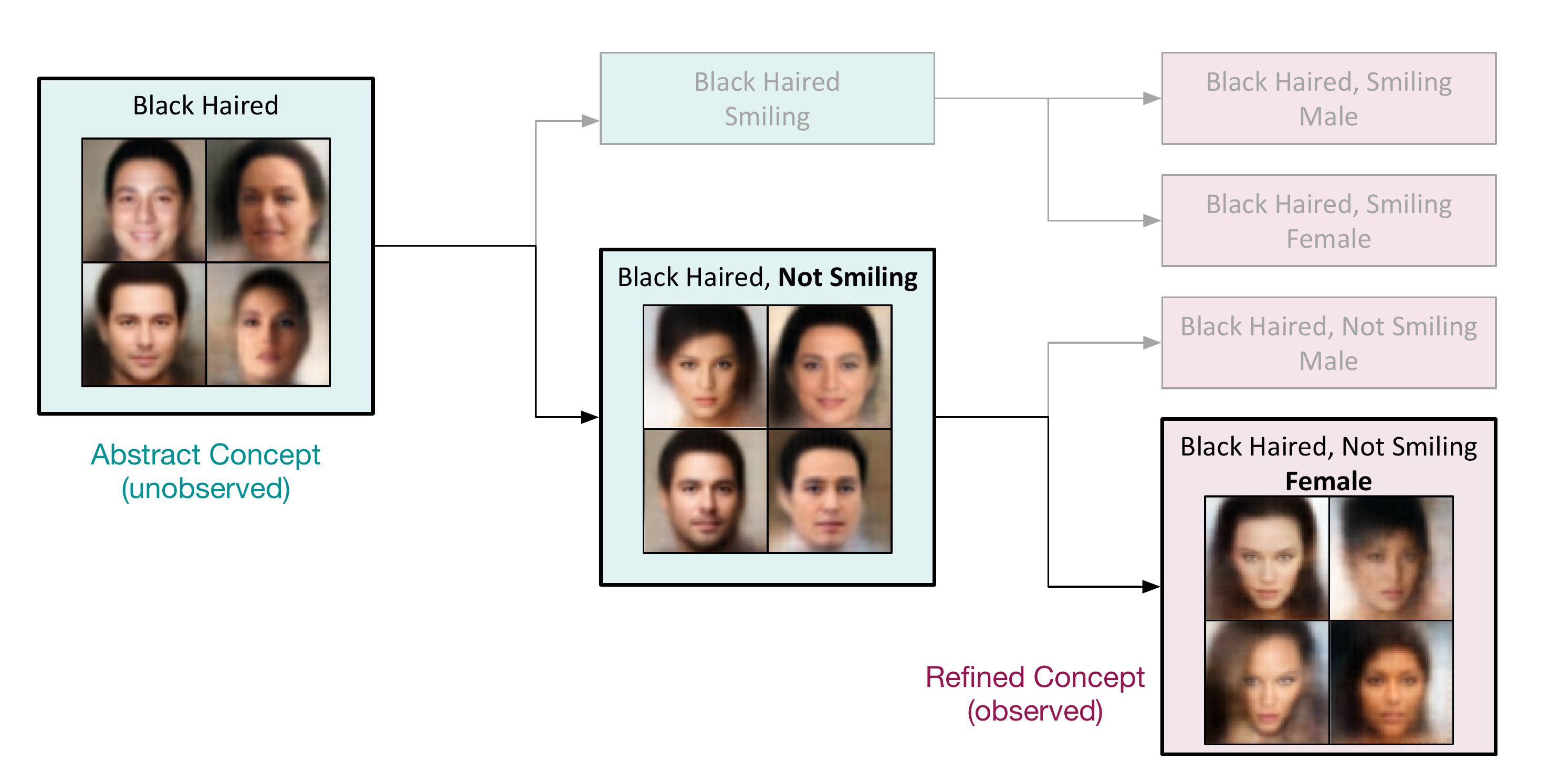}
	\caption{
          A compositional abstraction hierarchy  for faces,
          derived from 3 attributes:
          hair color, smiling or not, and gender.
          We show a set of sample images generated by our model,
          when trained on \CelebA,
          for different nodes in this hierarchy.
	}
	\label{fig:teaser}
	\vspace{-10pt}
\end{figure}

In this paper, we represent concept descriptions in terms of a
fixed length vector of discrete attributes $\attributes$.
This allows us to specify an exponentially large set of concepts
using a compact, combinatorial representation.
In particular, by specifying different subsets of attributes,
we can generate concepts at different levels of granularity or
abstraction.
We can arrange these concepts into
a {\em compositional abstraction hierarchy},
as shown in \cref{fig:teaser}.
\rama{This is a directed acyclic graph (DAG)
in which nodes represent concepts,
and an edge from a node to its parent
is added whenever we drop one of the attributes from the child's
concept definition. Note that we dont make any assumptions
	about the order in which the attributes are dropped (that is,
	dropping the attribute ``smiling'' is just as valid as
	dropping ``female'' in~\cref{fig:teaser}). Thus, the tree
	shown in the figure is just a subset extracted from the full
	DAG of concepts, shown for illustration purposes.}

We can describe a concept
by creating the attribute vector
$\vy_{\obs}$, in which we only specify the
value of the attributes in the subset $\obs \subseteq \attributes$;
the remaining attributes are unspecified, and are assumed to take all
possible legal values.
For example, consider the following concepts,
in order of increasing abstraction:
$C_{msb} = (\mathrm{male},\mathrm{smiling},\mathrm{blackhair})$,
$C_{*sb} = (*, \mathrm{smiling},\mathrm{blackhair})$,
and
$C_{**b} = (*,*,\mathrm{blackhair})$,
where the attributes are gender, smiling or not,
and hair color,
and $*$ represents ``don't care''.
A good model should
be able to generate images from different levels of the
abstraction hierarchy, as shown in \cref{fig:teaser}.
(This is in contrast to most prior work on conditional generative models of images,
which assume that all attributes are fully specified, which corresponds
to sampling only from leaf nodes in the hierarchy.)

In \cref{sec:methods}, we show how we can extend the
variational autoencoder (VAE) framework of \citet{VAE} to
create models
which can perform this task. The first extension
is to modify the model to the ``multi-modal'' setting
where we have both an image, $\vx$, and an attribute vector, $\vy$. More
precisely, we assume a joint generative model of the form $p(\vx,\vy,\vz) =
p(\vz) p(\vx|\vz) p(\vy|\vz)$, where $p(\vz)$ is the prior over the latent variable
$\vz$, $p(\vx|\vz)$ is our image decoder, and $p(\vy|\vz)$ is our description
decoder.
%This is the same model family as used in
%(see e.g., \citep{Wang2016,Pu2016,Yan2016,Pandey2017,Suzuki2017})
%various other papers (see e.g., %\citep{Wang2016,Pu2016,Suzuki2017}).
We additionally assume that the description decoder factorizes over
the specified attributes in the description,
so $p(\vy_{\obs}|\vz) = \prod_{k \in \obs} p(y_k|\vz)$.

We further extend the VAE
by devising a novel objective function,
which we call the \emph{\telbo}, for training the model
from paired data, $\data = \{ (\vx_n,\vy_n)\}$.
\eat{
but which can also handle unaligned data,
$\data_x = \{ \vx_n\}$
and
$\data_y = \{ \vy_n\}$.
}
However, at test time, we will allow unpaired data
(either just a description or just an image).
Hence we fit
three inference networks: $q(\vz|\vx,\vy)$, $q(\vz|\vx)$ and
$q(\vz|\vy)$.
This way we can embed an image or a description into the
same shared latent space (using $q(\vz|\vx)$ and $q(\vz|\vy)$, respectively);
this lets us ``translate'' images into descriptions or vice versa,
by computing $p(\vy|\vx) = \int d\vz \; p(\vy|\vz)
q(\vz|\vx)$ and $p(\vx|\vy) = \int d\vz \; p(\vx|\vz) q(\vz|\vy)$.
\eat{
Having the single modality inference networks,
$q(\vz|\vx)$ and $q(\vz|\vy)$,
also lets us learn from images without
descriptions, and descriptions without images, \ie, we can perform
semi-supervised learning.
}

To handle abstract concepts
(i.e., partially observed attribute vectors),
we use a method based on the
product of experts (POE)
\citep{Hinton2002}.
In particular, our inference network for attributes
has the  form $q(\vz|\vy_{\obs}) \propto p(\vz) \prod_{k \in \obs} q(\vz|\vy_k)$.
If no attributes are specified, the posterior is equal to the prior.
As we condition on more attributes, the posterior becomes narrower,
which corresponds to specifying a more precise concept.
This enables us to generate
a more diverse set of images to represent abstract concepts,
and a less diverse set of images to represent concrete concepts,
as we show below.

\cref{sec:eval} discusses how
to evaluate the performance of our method in an objective way.
Specifically, we first ``ground'' the description by generating a set of images,
$\calS(\vy_{\obs}) = \{ \vx^s \sim p(\vx|\vy_{\obs}): s=1:S \}$.
We then check that
all the sampled images in $\calS(\vy_{\obs})$
are consistent with the
specified attributes $\vy_{\obs}$
(we call this {\bf correctness}).
We also check that
the set of images ``spans'' the extension of the concept,
by exhibiting suitable diversity
(c.f. \citep{Young2014}).
Concretely, we check that the attributes that were {\em not specified}
(e.g., gender in $C_{*sb}$ above) vary across the different images;
we call this {\bf coverage}.
Finally,
we want
the set of images to have high
correctness and coverage even if the concept $\vy_{\obs}$
has a combination of attribute values that have not been seen in training.
For example, if we train on
$C_{msb} = (\mathrm{male},\mathrm{smiling},\mathrm{blackhair})$,
and
$C_{fnb} = (\mathrm{female},\mathrm{notsmiling},\mathrm{blackhair})$,
we should be able to test on
$C_{mnb} = (\mathrm{male},\mathrm{notsmiling},\mathrm{blackhair})$,
and
$C_{fsb} = (\mathrm{female},\mathrm{smiling},\mathrm{blackhair})$.
We will call this property {\bf compositionality}.
Being able to generate plausible
images in response to truly  compositionally novel queries
is the essence of imagination.
Together, we call these criteria
{\em the 3 C's of visual imagination}.

\cref{sec:results} reports experimental results
on two different datasets.
The first dataset is a modified version of MNIST, which we call
MNIST-with-attributes (or \MNISTa),
in which we
``render'' modified versions of a single MNIST digit on a
64x64 canvas, varying its location, orientation and size.
The second dataset is \CelebA\ \citep{CelebA},
which consists of over 200k face images,
annotated with 40 binary attributes.
We show that our method outperforms previous methods on these datasets.

The contributions of this paper are threefold.
First, we present a novel extension to VAEs in the multimodal setting,
introducing a principled new training objective (the \telbo), and deriving
an interpretation of a previously proposed objective (\jmvae)~\citep{Wang16ccacca} as a valid
alternative in \cref{sec:JMVAE}.
Second, we present a novel way to handle missing data in inference
networks based on a product of experts.
Third, we present novel criteria (the 3 C's) for evaluating
conditional generative
models of images, that extends prior work by considering the notion of visual
abstraction and imagination.

\section{Methods}
\label{sec:method}
\label{sec:methods}
\label{sec:models}

We start by describing standard VAEs,
to introduce notation. We then discuss our extensions to handle the
multimodal and the missing input settings.

\paragraph{Standard VAEs.}
A  variational autoencoder \citep{VAE} is a latent variable model of the form
$\pp(\vx,\vz) = \pp(\vz) \pp(\vx|\vz)$,
where $\pp(\vz)$ is the prior (we assume it is Gaussian,
$\pp(\vz)= \gauss(\vz|\vzero,\vI)$, although this assumption can be relaxed),
and $\pp(\vx|\vz)$ is the likelihood (sometimes called the
decoder), usually represented by a neural network.
To perform approximate posterior inference, we fit an inference
network (sometimes called the encoder) of the form $q_{\vphi}(\vz|\vx)$,
so as to maximize
$\calL(\vtheta,\vphi) =
\expectQ{\elbo(\vx,\vtheta,\vphi)}{\pdata(\vx)}$,
where
 $\pdata(\vx) = \frac{1}{N} \sum_{n=1}^N \delta_{\vx_n}(\vx)$ is the
empirical distribution,
and ELBO is the evidence lower bound:
\begin{equation}
  \elbo_{\lambda,\beta}(\vx,\vtheta,\vphi) =
  \expectQ{\lambda \log \pp(\vx | \vz)}
          {\qq(\vz|\vx,\vphi)}
          - \beta \KL( \qq(\vz| \vx), \pp(\vz) )
\end{equation}
Here $\KL(p,q)$ is the Kullback Leibler divergence between distributions
$p$ and $q$.
By default,  $\beta=\lambda=1$,
in which case we will
just write $\elbo(\vx,\vtheta,\vphi)$.
However, by using  $\beta>1$ we can encourage the posterior to be closer
to the factorial prior $p(\vz)=\gauss(\vz|\vzero,\vI)$,
which encouarges the latent factors to be ``disentangled'',
as proved in \citet{Achille2017};
this is known as the $\beta$-VAE trick \citep{Higgins2017}.
And allowing $\lambda>1$ will be useful later,
when we have multiple modalities.
%Due to the non-negativity of the $\KL$ divergence, $\elbo(\vx,\vtheta,\vphi) \leq \log
%\pp(\vx)$,
%so this  is a form of approximate maximum likelihood training.

\paragraph{Joint VAEs and the \telbo.}
\label{sec:elbo}

We extend the VAE to model images and attributes by defining
the joint distribution
$\pp(\vx,\vy,\vz) = \pp(\vz) \pp(\vx|\vz) \pp(\vy|\vz)$,
where $\pp(\vx|\vz)$ is the image decoder (we use the DCGAN
architecture from \citet{Radford2016}), and $\pp(\vy|\vz)$ is an MLP for
the attribute vector.
The corresponding training objective which we want to maximize becomes
$\calL(\vtheta,\vphi)=
\expectQ{\elbo(\vx,\vy,\vtheta,\vphi)}{\pdata(\vx,\vy)}$,
where
$\pdata(\vx,\vy) = \frac{1}{N} \sum_{n=1}^N \delta_{\vx_n}(\vx) \delta_{\vy_n}(\vy_n)$
  is the empirical distribution derived from
  paired data,
and the joint ELBO is given by
\begin{multline*}
\elbo_{\lambda_x,\lambda_y,\beta}(\vx,\vy,\vtheta_x,\vtheta_y,\vphi)=
\expectQ{
  \lambda_x \log \px(\vx|\vz)+\lambda_y \log \py(\vy|\vz)}{\qq(\vz|\vx,\vy)}
  \\
  - \beta \KL(\qq(\vz|\vx,\vy), \pp(\vz))
\end{multline*}
We call this the JVAE (joint VAE) model.
We usually set $\beta=1$,
but set  $\lambda_y/\lambda_x >1$ to 
to scale up
the likelihood from the low dimensional attribute vector,
 $\pp(\vy|\vz)$, to match the  likelihood from the high dimensional image,
$\pp(\vx|\vz)$.

Having fit the joint model above, we can proceed to train unpaired inference
networks $\qx(\vz|\vx)$ and $\qy(\vz|\vy)$, so we can embed images and
attributes into the same shared latent space.
Keeping the $p$ family fixed from the joint model, a natural objective to fit, say,
$\qx(\vz|\vx)$ is to maximize the following:\footnote{
  A reasonable alternative would be to minimize
  $\expectQ{\KL(\px(\vz|\vx), \qx(\vz|\vx))}{\pdata(\vx)}$.
  However, this is intractable, since we cannot compute
  $\px(\vz|\vx)$, by assumption.
  }
\begin{align*}
  \calL(\vphi_x|\vtheta)
&=  -\expectQ{\KL(\qx(\vz|\vx), \px(\vz|\vx))}{\pdata(\vx)} \\
  &= \int \int d\vx d\vz \; \pdata(\vx) \qx(\vz|\vx) \left[
    -\log \qx(\vz|\vx) - \log \px(\vx)
    + \log \px(\vx|\vz) + \log \pp(\vz) \right] \\
  &= \expectQ{\elbo(\vx,\vtheta_x,\vphi_x)}{\pdata(\vx)}
  - \expectQ{\log \px(\vx)}{\pdata(\vx)}
\end{align*}
where the last term is constant wrt $\vphi_x$ and the model family $p$,
and hence can be dropped.
We can use a similar method to fit $\qy(\vz|\vy)$.
Combining these gives  the following triple ELBO ({\em \telbo}) objective:
\begin{multline}\label{eqn:telbo}
  \calL(\vtheta_x,\vtheta_y,\vphi,\vphi_x,\vphi_y)
  =
  \mathbf{E}_{\hat{p}(\vx, \vy)}\left[
    \elbo_{1,\lambda,1}(\vx,\vy,\vtheta_x,\vtheta_y,\vphi)\right.\\
  \left. + \elbo_{1,1}(\vx,\vtheta_x,\vphi_{x})
  +
\elbo_{\gamma,1}(\vy,\vtheta_y,\vphi_{y})\right]
\end{multline}
%In practice, we also use a scaling factor $\beta_y$
%for the likelihood term in $\elbo(\vy)$ to balance the relative
%contribution of likelihoods across the unimodal ELBO terms.
%By default, we set $\beta_x=\beta_y=\beta_{xy}=1$,
%and tune $\lambda$ by cross validation.
where $\lambda$ and $\gamma$ scale the log likelihood
terms $\log p(\vy|\vz)$;
we set these parameters using a validation set.
Since we are training the generative model only on
aligned data, and simply retrofitting inference networks,
we freeze the $\px(\vx|\vz)$ and $\py(\vy|\vz)$ terms when
training the last two ELBO terms above,
and just optimize $\qx(\vz|\vx)$ and $\qy(\vz|\vy)$ terms.
This enables us to optimize all terms in~\cref{eqn:telbo} jointly.
Alternatively, we can first fit the joint model, and then
fit the unimodal inference networks.\footnote{
If we have unlabeled image data, we can perform semisupervised
learning
by optimizing $\expectQ{\elbo(\vx,\vtheta_x,\vphi_{x})}{\pdata(\vx)}$
wrt $\vtheta_x$ and $\vphi_x$,
as in \citet{Pu2016nips}.
}
In \cref{sec:related}, we compare this to other methods for training
joint VAEs that have been proposed in the literature.

\paragraph{Handling missing attributes.}
\label{sec:poe}
\label{sec:POE}
\begin{figure}
	\centering
	\includegraphics[page=1,width=\linewidth]{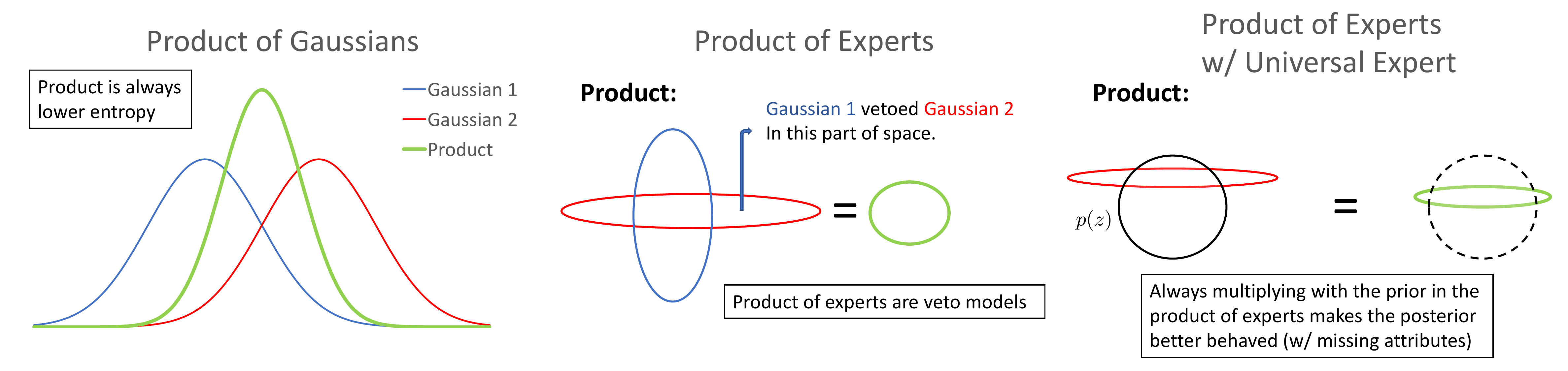}
	\caption{
          \small
          Illustration of the product of experts inference network.
          Each expert votes for a part of latent space
          implied by its observed attribute. The final posterior
          is the intersection of these regions.
          When all attributes are observed, the posterior will be a
          narrowly defined Gaussian, but when some attributes are
          missing,
          the posterior will be broader.
          Right: we illustrate how inclusion of
          the ``universal expert'' $p(\vz)$ in the product
          ensures that the posterior is always well-conditioned (close
          to spherical),
          even when we are missing some attributes.
          }
        	\label{fig:poe_intuition}
        	\vspace{-10pt}
\end{figure}
\rama{In order to handle missing attributes at test time, we use
a product of experts model, where each attribute instantiates an
expert. We are motivated by prior work~\citep{Williams18} which shows that for
a linear factor analysis model, the posterior distribution $p(\vz|\vy)$ is a product
of $K$-dimensional Gaussians, one for each visible dimension.
 Since our model is just a nonlinear extension of factor analysis, we choose the form of the approximate posterior of our inference network, $q(\vz|\vy)$, to be a product of Gaussians, one for each visible feature:
 $q(\vz|\yobs) \propto p(\vz) \prod_{k \in \obs} q(\vz|y_k)$,
where  $q(\vz|y_k) = \gauss(\vz
| \vmu_k(y_k),\vC_k(y_k))$ is the $k$th Gaussian ``expert'',
and $p(\vz) = \gauss(\vz|\vmu_0=\vzero, \vC_0=\vI)$ is the prior.
A similar model was concurrently proposed in~\citet{Bouchacourt2018} to perform inference
for a set of images.
Unlike
the product of experts model in \citep{Hinton2002},
our model multiplies Gaussians, not Bernoullis,
so the product has a closed form solution
namely
$q(\vz|\yobs) = \gauss(\vz | \vmu, \vC)$,
where $\vC^{-1} = \sum_{k} \vC_k^{-1}$ and
$\vmu = \vC(\sum_k \vC_k^{-1} \vmu_k)$,
and the sum is over all the observed attributes.}
\rama{Intuitively, $\vy$ imposes an increasing number of constraints on $\vz$ as more of it is observed,
as explained in \cite{Williams02}.
In our setting, if we do not observe any attributes, the posterior reduces to the
prior. As we observe more attributes, the posterior becomes narrower,
since the (positive definite) precision matrices, $\vC^{-1}$ add up,
reflecting the increased specificity of the concept being specified,
as illustrated in \cref{fig:poe_intuition} (middle) (see also~\cite{Williams02}).
We always include the prior term, $p(\vz)$, in the product,
since without it, the posterior $\qy(\vz|\yobs)$ may not be well-conditioned
when we are missing attributes,
as illustrated in \cref{fig:poe_intuition} (right).
For more implementation-level details on the model architectures,
see \cref{sec:arch}.}

\eat{
This method differs from the product of experts (POE)
model proposed in \citet{Hinton2002} in several ways:
we use it represent the distribution of the
latent variables, not the visible variables;
we use products of Gaussians, not products of Bernoullis,
which makes all the computations analytically tractable;
and we include the ``universal expert'' $p(\vz)$.
To see why we need the universal expert,
note that when all attributes are present
(as is the case during training),
it is okay
if individual Gaussian experts are  very  broad in some
latent dimensions,
as long as the overall product is ``well behaved''.
However, at test time, some attributes may be missing,
so the ``censoring'' by certain experts  is turned off,
so
$q(\vz|\yobs)$ can correspond to a Gaussian with
a very poorly conditioned covariance matrix, resulting
in a very long, thin distribution along certain dimensions.
By always including
$p(\vz)$ in the product,  we ensure that the resulting
posterior is always well conditioned,
which results in improved results.
}

\section{Evaluation metrics: The 3C's of Visual Imagination}
\label{sec:metrics}
\label{sec:eval}

\eat{
To assess the quality of a conditional generative model of images,
we draw inspiration from the field of education, which similarly faces
the challenge of assessing whether a student has successfully
``understood'' a concept (\cf, \citep{Piech2015}).
With visual concepts, a natural approach is to give the student a
description of the concept, and ask them to generate $N$ images that match
that description,
which we denote by
$\extension(\concept) = \{ \vx_n \sim p(\vx|\concept): n=1:N \}$.
}

To evaluate the quality of a set of generated images,
$\extension(\concept) = \{ \vx_s \sim p(\vx|\concept): s=1:S \}$,
we apply a multi-label classifier to each image,
to convert it to a predicted attribute vector,
$\hat{\vy}(\vx)$.
This attribute classifier  is  trained
on a large dataset of images and
attributes, and is held constant across all methods that are being evaluated.
It plays the role of a human observer.
This is similar in spirit to generative
adversarial networks \citep{GAN}, that declare a generated
image to be good enough if a binary classifier cannot distinguish it
from a real image. (Both approaches avoid the problems mentioned
in \citet{Theis2016} related to evaluating generative image models in
terms of their likelihood.)
However, the attribute classifier checks not only
that the images look realistic, but also that they have the desired attributes.

To quantify this,
we define the {\bf correctness} as the fraction of attributes for each generated image
that match those specified in the concept's description:
%
%{\footnotesize
%\begin{equation}
$\mathrm{\correctness}(\extension, \concept) =
\frac{1}{|\extension|}
\sum_{\vx \in \extension} \frac{1}{|\obs|} \sum_{k \in \obs}
\ind{\hat{y}(\vx)_k = y_k}.
$
%\end{equation}
%}
%
However, we also want to measure the diversity of values
for the \emph{unspecified} or missing attributes,
$\missing = \attributes \setminus \obs$.
We do this by comparing  $q_k$,
the empirical distribution over values for attribute $k$
induced by the generated set $\extension$,
to $p_k$, the true distribution for this attribute
induced by the training set.
\eat{
  footnote{
  This ignores the fact that there might be interactions between the
  attributes.
  For example, there might not be any females with moustaches in the
  dataset, but the marginal distribution for moustaches might not
  indicate this.
  More precisely, let $k$ be the index for the moustache attribute,
  and assume it has two values, present and absent.
  If 40\% of the training images are men with moustaches,
  and 50\% of the training images are male,
  we will find $p_k=[0.2,0.8]$.
  But  if we generate images for the concept ``female'',
it may fail to add a moustache to any of these samples,
in which case we will estimate $q_k=[0.0, 1.0]$.
\TODO{Is this a fatal flaw?}
  }
  }
We measure the difference between these distributions
using the Jensen-Shannon divergence,
since it is symmetric and satisfies $0 \leq \JS(p,q) \leq 1$.
We then define
the {\bf coverage} as follows:
%
%{\footnotesize
%\begin{equation}
$\mathrm{coverage}(S, \concept) =
\frac{1}{|\missing|} \sum_{k \in \missing}
(1-\JS(p_k, q_k)).
$
%\end{equation}
%}
%
If desired, we can combine correctness and coverage into a single number,
by computing the JS divergence between
$p_k$ and $q_k$ for all attributes,
where, for observed attributes,
$p_k$ is a delta function
and $q_k$ is the empirical distribution \rama{(we call this \textbf{JS-overall}). This gives us a convenient
way to pick hyperparameters. However, for analysis, we find it helpful to report correctness and coverage separately.}

Note that our metric is different from the inception score
proposed in \citet{Salimans2016}.
That is defined as follows:
$  \mathrm{inception}
  = \exp\left(
  \expectQ{\KL(p(y|\vx), p(y))}{\pdata(\vx)} \right)$,
  where $y$ is a class label.
Expanding
  the term inside the exponential, we get
  \[
  \sum_{\vx} p(\vx)
  \left[ \sum_y p(y|\vx) \log p(y|\vx) \right]
  - \sum_{\vx} \sum_y p(\vx,y) \log p(y)
  = \expectQ{-H(y|\vx)}{\pdata(\vx)}
  +H(y)
  \]
A high inception score means that the distribution
$p(y|\vx)$ has low entropy, so the generated images match some class,
but that the marginal $p(y)$ has high entropy,
so the images are diverse.
However, the inception score was created to evaluate
unconditional generative models of images,
so it does not check if the generated images are consistent with
the concept $\vy_{\obs}$,
and the degree of diversity does not vary in
response to the level of abstraction of the concept.

 Finally, we can assess how well the model understands
 {\bf compositionality},
 by checking correctness of its generated images
 in response to test concepts $\vy_{\obs}$
that differ in at least one attribute from the training concepts.
We call this a {\em compositional split} of the data.
This is much harder than a standard {\em iid} split,
since we are asking the model to predict the effects of novel
combinations of attributes, which it has not seen before
(and which might actually be impossible).
Note that abstraction is different from compositionality -- in
abstraction we are asking the model to predict the effects of
dropping certain attributes instead of predicting novel combinations of
attributes.
%For this reason, we only create a compositional split
%for the \MNISTA\ dataset.
%Note that we only measure correctness on the compositional split,
%since coverage is not well defined when all attributes are specified.

\eat{
By contrast, an iid split (e.g., as used in \CelebA)
allows (or even requires) the same labels
to occur in the training and test split (although the images will not overlap).
}

\section{Related Work}
\label{sec:related}

In this section, we briefly mention some of the most closely related prior work.

\paragraph{Conditional models.}%
Many conditional generative image models of the form
$p(\vy|\vx)$ have been proposed recently,
where $\vy$ can be
a class label (e.g., \citep{Radford2016}),
a vector of attributes (e.g., \citep{Yan2016}),
a sentence (e.g., \citep{Reed2016generative}),
another image (e.g., \citep{pix2pix}),
etc.
Such models are usually based on VAEs or GANs.
However, we are more interested in learning a shared latent space from
either descriptions $\vy$ or images $\vx$,
which means we need to use a joint, symmetric, model.

\eat{
Conditional VAEs
 (\eg, \citep{Yan2016,Pandey2017})
 and conditional GANs
 (\eg, \citep{Reed_2016_ICML,Mansimov_2016_ICLR})
learn a stochastic mapping $p(\vx|\vy)$
from attributes $\vy$ to images $\vx$.
However, they cannot
 compute both $p(\vx|\vy)$ and $p(\vy|\vx)$, and do not support
 semi-supervised learning,  unlike our joint method.
More importantly,
these conditional models cannot handle missing inputs,
so they cannot be used to generate abstract visual concepts.
}

\paragraph{Joint models.}%
Several papers use the same joint VAE model as us,
but they differ in how it is trained.
In particular,  the \bivcca objective of \citet{Wang2016}
has the form
$\calL(\vtheta,\vphi) = \expectQ{J(\vx,\vy,\vtheta,\vphi)}{\pdata(\vx,\vy)}$,
where
%% \begin{align*}
%%   J(\vx,\vy,\vtheta,\vphi)=
%% \mu \left(
%% E_{\qx(\vz|\vx)}[ \log \pp(\vx,\vy|\vz)] - \KL(\qx(\vz|\vx), \pp(\vz)) \right)\\
%% + (1-\mu) \left( E_{\qy(\vz|\vy)} [\log \pp(\vx,\vy|\vz) ] - \KL(\qy(\vz|\vy), \pp(\vz)) \right)
%% \end{align*}
\begin{align*}
  J(\vx,\vy,\vtheta,\vphi)=
\mu \left(
E_{\qx(\vz|\vx)}[ \log \px(\vx|\vz) + \lambda \log \py(\vy|\vz)]
- \KL(\qx(\vz|\vx), \pp(\vz)) \right)\\
+ (1-\mu) \left( E_{\qy(\vz|\vy)} [\log \px(\vx|\vz) + \lambda \log
  \py(\vy|\vz) ]
- \KL(\qy(\vz|\vy), \pp(\vz)) \right)
\end{align*}
This method results in the model generating
the mean image corresponding to each concept,
due to the $E_{\qy(\vz|\vy)} \log \pp(\vx,\vy|\vz)$ term,
which requires that $\vz$'s sampled from
$\qy(\vz|\vy_n)$ be good at generating all the
different $\vx_n$'s which co-occur with $\vy_n$.
We show this empirically in \cref{sec:results}.
 This problem can be partially compensated for by increasing $\mu$,
 but that reduces the  $\KL(\qq(\vz|\vy), \pp(\vz))$ penalty,
 which is required to ensure $\qy(\vz|\vy)$ is a broad distribution
 with good coverage of the concept.

The JMVAE objective of \citet{Suzuki2017} has the form
 $\calL(\vtheta,\vphi) = \expectQ{J(\vx,\vy,\vtheta,\vphi)}{\pdata(\vx,\vy)}$,
 where
\[
J(\vx,\vy,\vtheta,\vphi)=
\elbo_{1,\lambda,1}(\vx,\vy,\vtheta,\vphi)
-\alpha \left[ \KL(\qq(\vz|\vx,\vy), \qy(\vz|\vy))
+ \KL(\qq(\vz|\vx,\vy), \qx(\vz|\vx)) \right]
\]
\eat{
Note that
$\qq(\vz|\vx,\vy) \approx \qx(\vz|\vx)$
when $(\vx,\vy)$ is a matching pair of images and labels,
since the image essentially uniquely determines the posterior
embedding;
hence the second KL term is unnecessary.
Furthermore, the first KL term looks rather odd,
}
At first glance, 
forcing $\qq(\vz|\vy)$ to be close
to $\qq(\vz|\vx,\vy)$ seems undesirable,
since the latter will typically be close to a delta function,
since there is little posterior uncertainty in $\vz$
once we see the image $\vx$.
However,
in \cref{sec:JMVAE},
we use results from \citet{elboSurgery} to show
%we show
that $\expectQ{\KL(\qq(\vz|\vx,\vy), \qy(\vz|\vy))}{\pdata(\vx,\vy)}$
can be written in terms of
$\KL(\qqavg(\vz|\vy), \qy(\vz|\vy))$,
where $\qqavg(\vz|\vy) = \expectQ{\qq(\vz|\vx,\vy)}{\pdata(\vx|\vy)}$
is the aggregated posterior over $\vz$ induced by all images
$\vx$ which are associated with description $\vy$.
This ensures that $\qy(\vz|\vy)$ will cover
the embeddings of {\em all} the images associated
with concept $\vy$.
However, since there is no $\KL(\qy(\vz|\vy),\pp(\vz))$
term,
the diversity of the samples is slightly reduced
for novel concepts compared to
\telbo, as we show empirically in \cref{sec:results}.
\rama{On the flip side, the benefit of using the aggregated
posterior to fit the $q(\vz|\vy)$ inference network is that one can expect
sharper images, as this ensures we will sample $\vz \sim q(\vz|\vy)$ which have been
seen by
the image decoder $\pp(\vx| \vz)$ during joint training.
If the aggregated posterior does not exactly match the prior (which is known to happen in VAE-type models, see~\cite{elboSurgery})
then regularizing with respect to the prior (as \telbo does) can generate samples in parts
of space not seen by the image decoder, which can potentially lead to less ``correct''
samples. Again, our empirical findings in~\cref{sec:results} confirm
this tradeoff between correctness and coverage implicit in choices of
\telbo \vs{} \jmvae.}

\eat{
This is potentially problematic,
since the $\KL(q(\vz|\vx,\vy), \rama{r(\vz|\vy)})$ penalty
may result in lack of coverage of a concept,
because $q(\vz|\vy)$ should be a broad distribution,
to cover all the variations of the concept,
whereas $q(\vz|\vx,\vy)$ will typically be close to a
delta function,  because there is usually little posterior uncertainty about the latent
factors given an image.
By contrast, in our \telbo\ objective, we minimize
$\KL(q(\vz|\vy), p(\vz|\vy)$,
which tries to approximate the true (broad) posterior.
The method of \citep{Pu2016nips} is similar to ours,
but focuses on the $p(\vy|\vx)$ mapping, and does not fit a $q(\vz|\vy)$ network.
}

The SCAN method of \citet{Higgins2017scan}
first fits a standard
$\beta$-VAE model \citep{Higgins2017}
on unlabeled images
(or rather, features derived from images using a pre-trained denoising
autoencoder)
by maximizing
$\calL(\vtheta_x,\vphi_x) = \expectQ{\elbo_{1,\beta_x}(\vx,\vtheta_x,\vphi_x)}{\pdata(\vx)}$.
They then 
 fit a second VAE by maximizing
$\calL(\vtheta_y,\vphi_y) = \expectQ{J(\vx,\vy,\vtheta_y,\vphi_y,\vphi_x)}{\pdata(\vx,\vy)}$,
where
\[
J(\vx,\vy,\vtheta_y,\vphi_y,\vphi_x)=
\elbo_{1,\beta_y}(\vy,\vtheta_y,\vphi_y)
-\alpha \KL(\qx(\vz|\vx), \qy(\vz|\vy))
\]
This is very similar to JMVAE,
since $\qx(\vz|\vx) \approx \qq(\vz|\vx,\vy)$,
when $(\vx,\vy)$ is a matching pair of images and labels.
An important difference, however,
is that SCAN treats the attribute vectors $\vy$ as
atomic symbols;
this has the advantage that there is no need to handle missing inputs,
but the disadvantage that
they cannot infer the meaning of unseen
attribute combinations at test time, unless they are ``taught''
them by having them paired with images.
Also, they rely
on $\beta_x > 1$
as a way to get compositionality, assuming that a disentangled
latent space will suffice.
However, in \cref{sec:mnist-latent}, we show that
unsupervised learning of the latent space
given images alone can result in poor
results when some of the attributes in the compositional concept hierarchy
are non-visual,
such as parity of an MNIST digit.
Our approach always takes the labels into consideration when learning
the latent space, permitting well-organized latent spaces even in the
presence of non-visual concepts
(c.f. the difference between PCA and LDA).

\paragraph{Handling missing inputs.}%
Conditional generative models of images, of the form
$p(\vx|\vy)$,
have problems with missing input attributes,
as do inference networks $q(\vz|\vy)$ for VAEs.
 \citet{Hoffman2017} uses MCMC to fit a latent Gaussian model,
 which can in principle handle missing data;
 however, he initializes the Markov chain
 with the posterior mode computed by an inference network,
 which cannot easily handle missing inputs.
One approach we can use,
if we have a joint model,
is to  estimate or impute the missing values,
as follows:
$\hat{\vy} = \arg \max_{\ymiss} p(\ymiss|\yobs)$,
where $p(\ymiss,\yobs)$ models dependencies between attributes.
We can then sample images using
$p(\vx|\hat{\vy})$.
This approach was used in \citet{Yan2016}
to handle the case where some of the pixels being passed into an
inference network were not observed.
However, conditioning on an imputed value will give different
results from not conditioning on the missing inputs; only the latter will increase the
 posterior uncertainty in order to correctly represent less precise
 concepts with broader support.

\paragraph{Gaussian embeddings.}
%\citep{Vilnis2015,Athiwaratkun2017,Mukherjee2016,Ren2016}
%learn Gaussian distributions for words and images.
%However, they train using a contrastive loss, based on similar and dissimilar words.
%In addition, the covariance matrix of their Gaussians is fixed, whereas ours vary dynamically depending on how many attributes we condition on.
There are many papers that embed images and text into points in a vector space.
However, we want to represent concepts of different levels of abstraction,
and therefore want to map images and text to regions of latent space.
There are some prior works  that use Gaussian embeddings for words
\citep{Vilnis2015,Athiwaratkun2017},
sometimes in conjunction with  images
\citep{Mukherjee2016,Ren2016}.
Our method
differs from these approaches
in several ways.
First, we maximize the likelihood of $(\vx,\vy)$ pairs,
whereas the above methods  learn a
Gaussian embedding using a contrastive loss.
Second, our PoE formulation ensures that the covariance
of the posterior $q(\vz|\yobs)$ is adaptive to the data that we condition on.
In particular, it becomes narrower as we observe more attributes (because the precision matrices sum up),
which is a property not shared by other embedding methods.

\paragraph{Abstraction and compositionality.}
\citet{Young2014} represent the extension of a concept (described by a
noun phrase)
in terms of a set of images whose captions match the phrase. By
contrast, we use a parametric probability distribution in a latent
space that can generate new images.
\citet{Vendrov2016} use order embeddings, where they explicitly learn
subsumption-like relationships by learning a space that respects
a partial order. In contrast, we reason about generality of concepts via
the uncertainty induced by their latent representation.
There has been some work
on compositionality in the language/vision
literature
(see e.g., \citet{Atzmon2016,Johnson2017,Agarwal_2017_arxiv}),
but none of these papers use generative models,
which is arguably a much more stringent test of whether a model
has truly ``understood'' the meaning of the components
which are being composed.

\section{Experimental results}
\label{sec:results}
\label{sec:experiments}

In this section, we fit the JVAE model to two different datasets
(\MNISTA and \CelebA),
using the \telbo\ objective, as well as \bivcca\ and \jmvae.
We measure the quality of the resulting model using the 3 C's,
and
show that our method of handling missing data behaves in a qualitatively reasonable way.

\subsection{\MNISTA}

\paragraph{Dataset.}
In this section, we report results on the \MNISTA\ dataset.
This is created by modifying the original MNIST dataset as follows.
We first create  a compositional concept hierarchy using
4 discrete
attributes, corresponding to class label (10 values),
location (4 values),
orientation (3 values), and size (2 values).
Thus there are \texttt{10x2x3x4=240} unique concepts in total.
We then sample $\sim 290$ example images of each concept, and create
both an iid and compositional split of the data.
See \cref{sec:mnistDetails} for details.

\paragraph{Models and algorithms.}
We train the JVAE model on this dataset using \telbo, \bivcca\ and
\jmvae\ objectives.
We use Adam~\citep{Kingma2015adam} for
	optimization, with a learning rate of 0.0001, and a minibatch size of 64.
	We train all models for 250,000
	steps (we generally found that the models do not tend to overfit in our
	experiments). \rama{Our models typically take around a day to train
	on NVIDIA Titan X GPUs.}
For the image models, $p(\vx|\vz)$ and $q(\vz|\vx)$,
we  use the DCGAN architecture from \cite{Radford2016}.
\rama{Our generated images are of size 64$\times$64, as in~\citet{Radford2016}.}
For the attribute models,  $p(y_k|\vz)$ and $q(\vz|y_k)$,
we use MLPs.
For the joint inference network, $q(\vz|\vx,\vy)$, we use a CNN
combined with an MLP.
We use $d=10$ latent dimensions for all models.
\rama{We choose the hyperparameters for each method
so as to maximize JS-overall, which is an overall
	measure of \correctness and \coverage (see \cref{sec:eval}) on a validation set of attribute queries.}
See \cref{sec:arch}  for further details on the model architectures.

\paragraph{Evaluation.}
To measure correctness and coverage,
we first
train the observation classifier on the full \iid dataset,
where it gets to an accuracy of 91.18\% for
class label, 90.56\% for scale, 92.23\% for orientation, and 100\% for location.
Consequently, it is a reliable way to assess the quality of samples
from various generative models
(see \cref{sec:obsClassifier} for details).
We then compute correctness and coverage on the \iid dataset,
and coverage on the  \comp dataset.

\begin{figure}[t]
  \centering
	\includegraphics[width=\linewidth]{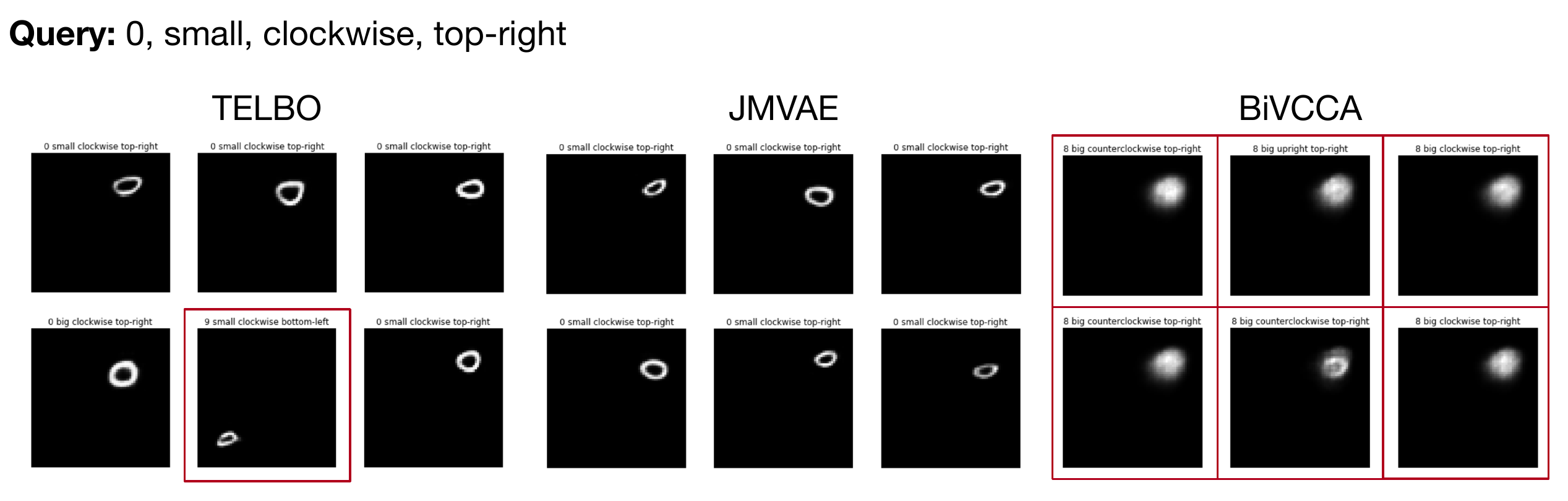}
	\caption{
          \small
          Samples from attribute vectors seen at training time,
          generated by the 3 different models.
          We plot the posterior mean of each pixel, $\expect{\vx|\vz_s}$,
          where $\vz_s \sim \qy(\vz|\vy)$.
          The caption at the top of each little image  is the
          predicted attribute values.
          The border of the generated image is red if any of the
          attributes are predicted incorrectly.
         (The observation classifier is fed sampled images, not the
          mean image that we are showing here.)
		}
	\label{fig:concreteSamples}
	\vspace{-10pt}
  \end{figure}

\begin{figure}[ht]
  \centering
  \begin{tabular}[c]{c c}
      \begin{subfigure}[c]{0.4\textwidth}
    \centering
\includegraphics[width=\linewidth]{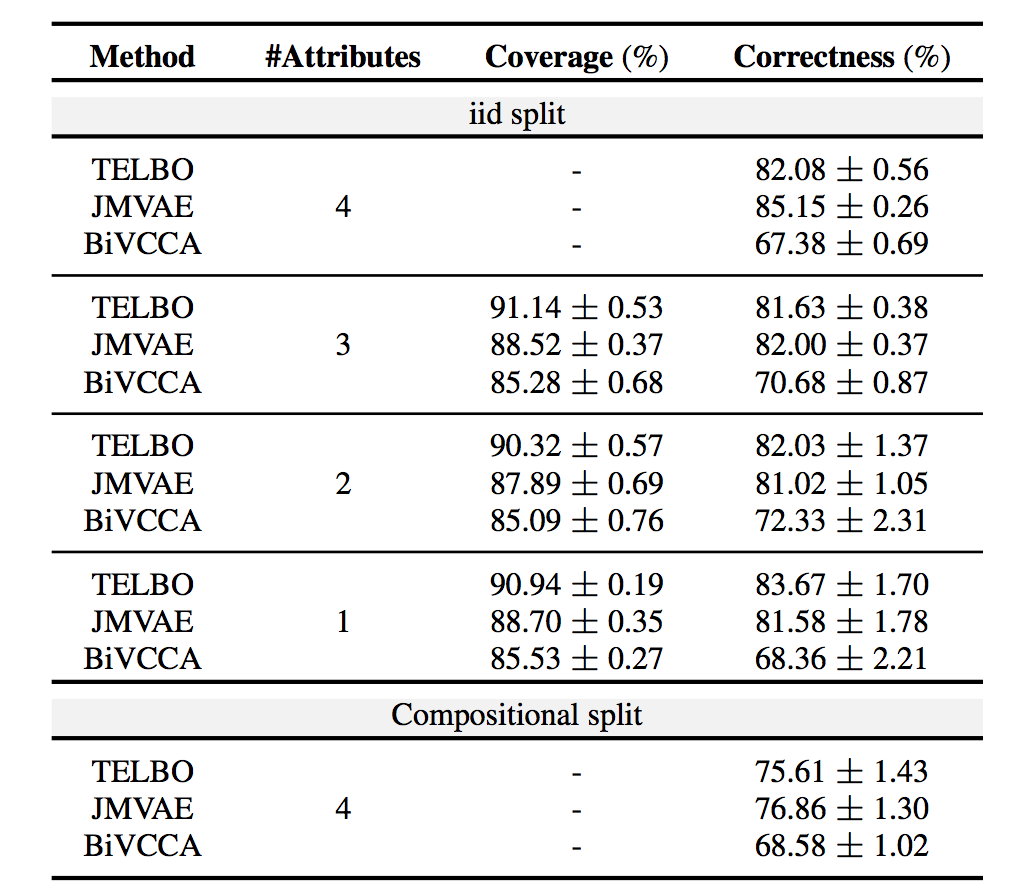}
\caption{
  \small
		Evaluation of different approaches on the
		test set. Higher numbers are better. We report standard deviation
		across 5 splits of the test set.
	}
	\label{tab:metrics}
\end{subfigure}&
    \begin{subfigure}[c]{0.55\textwidth}
    \centering
	\includegraphics[width=1.0\linewidth]{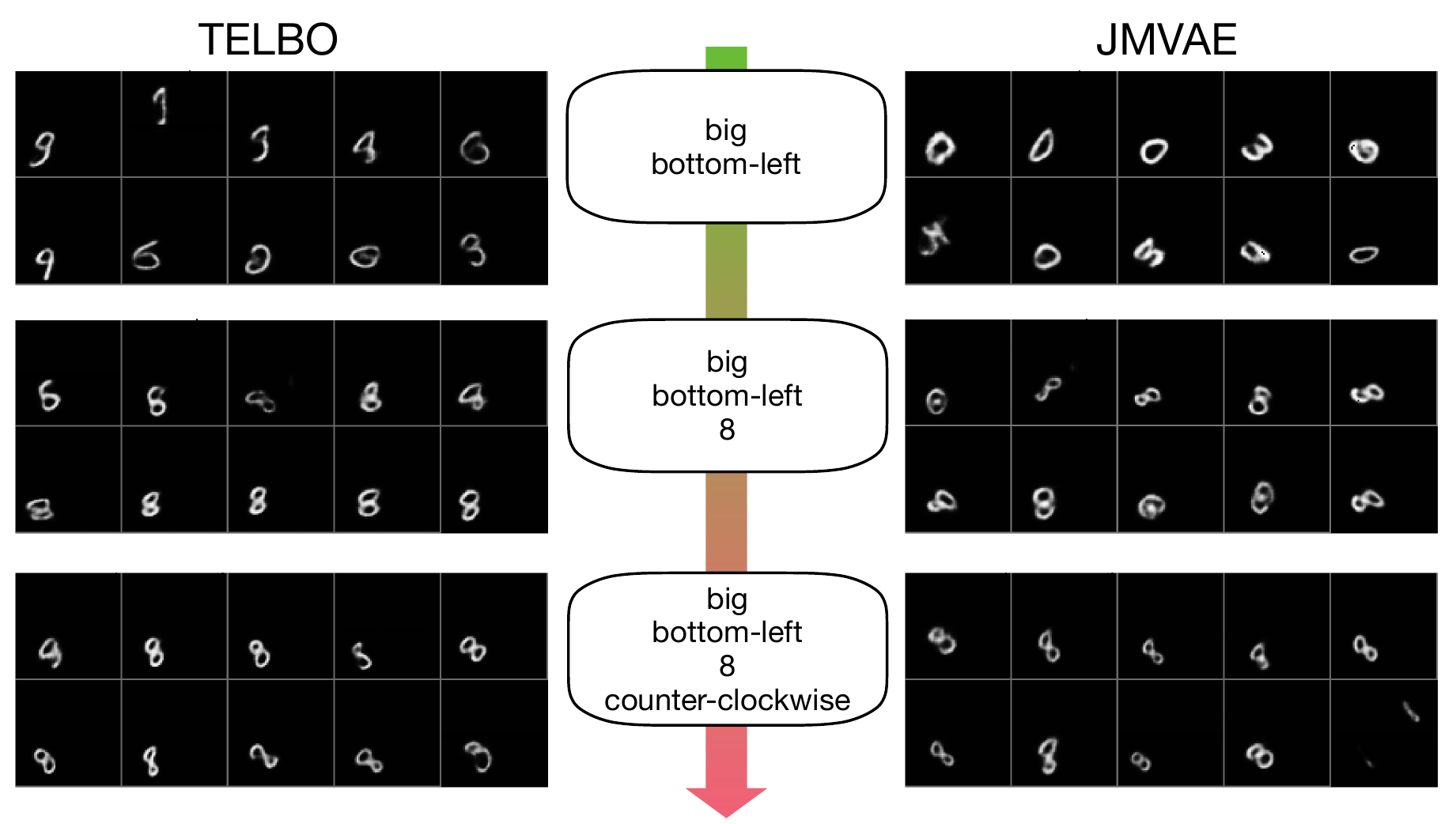}
	\caption{
          \small
		Mean images generated by \telbo and \jmvae in response
		to queries at different levels of abstraction, starting from
		abstract (top) to refined (bottom).
	}
	\label{fig:abstractHierarchyMnist}
    \end{subfigure}
\end{tabular}
    \caption{\textbf{(a)} We show quantitaive results on the 3C's on \MNISTA. \textbf{(b)} Qualitative results on \MNISTA for various queries. For refined/fully specified queries, we can see that both \telbo and \jmvae produce
    		good correctness, \ie, the images produced follow
    		constraints placed by the specified attributes. When the attribute
    		`orientation' is unspecified, we see that \telbo produces upright
    		and counter clockwise digits, while \jmvae produces clockwise
    		and upright digits. Finally, when we leave the digit unspecified
    		(top), we see that \telbo appears to generate a more diverse set
    		of digits (9, 3, 8, 6) while \jmvae produces 0 and 3.}
    \label{fig:mnistResults}
    \vspace{-20pt}
\end{figure}

\paragraph{Familiar concrete concepts.}%
We start by assessing the quality of the models in the simplest
setting, which is where the test concepts are fully specified (i.e.,
all attributes are known), and the concepts have been seen before in
the training set (i.e., we are using the \iid\ split).
\cref{tab:metrics}
%\cref{fig:mnist-results}
shows the correctness scores for the three methods.
(Since the test concepts are fully grounded, coverage is not well
defined, since there are no missing attributes.)
We see that \telbo has a correctness of 82.08\%,
which is close to that of \jmvae (85.15\%);
both methods significantly outperform
\bivcca (67.38\%).
To gain more insight,
\cref{fig:concreteSamples} shows some samples
from each of these methods for a leaf concept
chosen at random.
We see that the images generated by \bivcca are very blurry,
for reasons we discussed in \cref{sec:related}.
Note that these blurry images are correctly detected
by the attribute classifier.\footnote{
  We chose the value  of $\mu=0.7$ based on
  maximizing correctness score on the validation set.
Nevertheless, this does not completely eliminate blurriness, as we can see.
}
%The blurriness is
%because of the $E_{\qy(\vz|\vy)}[\log \pp(\vx,\vy|\vz)]$ term,
%as we discussed .
We also see that the \jmvae samples all look
good (in this example).
Most of the samples from \telbo\ are also good,
although there is one error
(correctly detected by the attribute classifier).

\paragraph{Novel abstract concepts.} %
Next we assess the quality of the models when the test concepts are
abstract, i.e., one or more attributes are not specified.
(Note that the model was never trained on such  abstract
concepts.)
\cref{tab:metrics}
shows that the correctness scores for
\jmvae\ seems to drop somewhat
(from about 85\% to about 81.5\%),
although it remains steady for \telbo\ and \bivcca.
We also see that the coverage of \telbo\ is higher
than the other methods,
due to the use of the $\KL(\qy(\vz|\vy),\pp(\vz))$ regularizer,
as we discussed in \cref{sec:related}.
\cref{fig:abstractHierarchyMnist}
illustrates how the methods respond to concepts
of different levels of abstraction.
The samples from the \telbo\ seem to be more diverse,
which is consistent with the numbers in
\cref{tab:metrics}.

\paragraph{Compositionally novel concrete concepts.}%
Finally we assess the quality of the models when the test concepts are
fully specified, but have not been seen before (i.e., we are using the
\comp\ split).
\cref{tab:metrics} shows some quantitative results.
We see that the correctness for \telbo and \jmvae
has dropped from about 82\% to about 75\%,
since this task is much harder, and requires
``strong generalization''.
However, as before, we see that both \telbo and \jmvae
outperform \bivcca,
which has a correctness of about 69\%. See \cref{subsec:comp_generalization_mnista} qualitative results and more details.

\subsection{\CelebA}
\label{sec:celebAresults}

\begin{figure}[ht]
  \centering
  %  \fbox{\rule[-.5cm]{0cm}{4cm} \rule[-.5cm]{4cm}{0cm}}
  	\includegraphics[width=\linewidth]{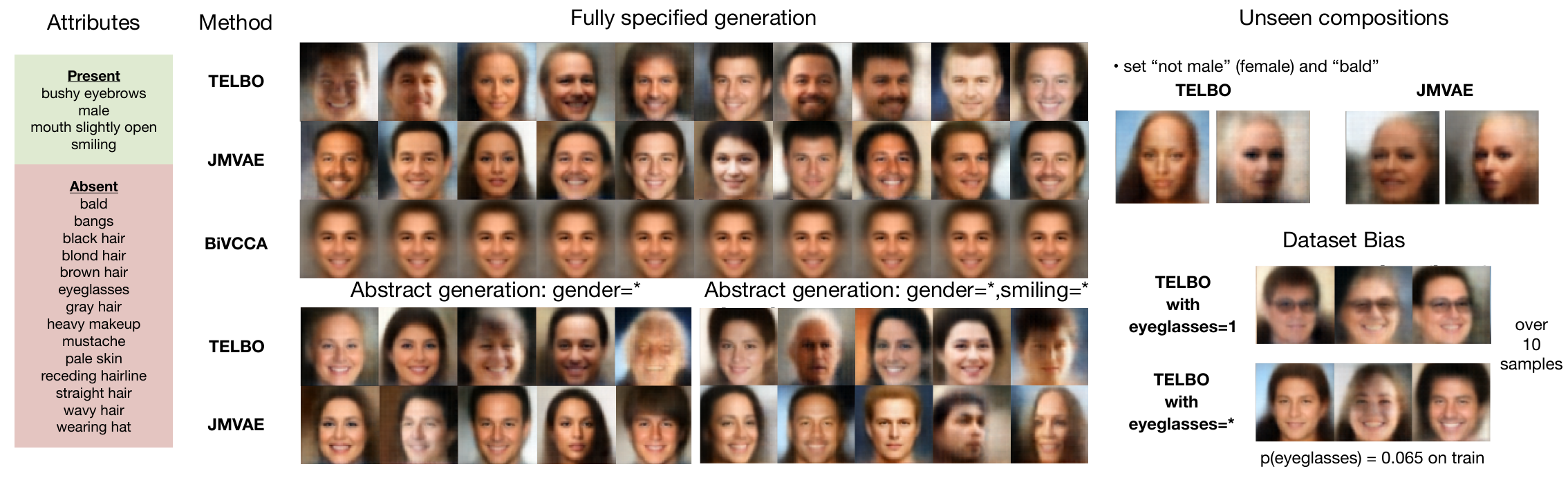}
  \caption{
    Sample \CelebA\ results. Left: we show the attributes specified to be
    present or absent when generating images. Middle: we show 10 samples each generated from \telbo, \jmvae
    and \bivcca. We see that \telbo and \jmvae genreate better samples than \bivcca
    which collapses to the mean. Middle, bottom: We show five samples from \telbo
    and \jmvae in response to queries with unspecified attributes, and see that both
    approaches generate a mix in the samples, generalizing meaningfully across unspecified attributes.
	}
	\label{fig:celeba}
\end{figure}

In this section, we report results on the \CelebA\ dataset
\citep{CelebA}.
In particular,
we use the version that was used in \citet{Perarnau2016},
which selects 18 visually distinctive attributes, and generate
images of size 64$\times$64;
see \cref{sec:celebAdetails} for more details on the \CELEBA dataset
and~\cref{sec:arch} for details of the model architectures.
\cref{fig:celeba} shows some sample qualitative results.
On the top left, we show some images which were generated
by the three methods given the concept shown
in the left column.
\telbo and \jmvae generate realistic and diverse images. \rama{
That is, the generated images are generally of males, with
mouth slightly open and smiling attributes present in the images.
On the other hand, \bivcca just generates the mean image.
On the bottom left, we show what happens when we drop
some attributes, thus specifying more abstract concepts. 
We see that when we drop the gender, we get a mixture of
both male and female images for both \telbo and \jmvae. Going further,
when we drop the ``smiling'' attribute, we see that the samples
now comprise of people who are smiling as well as not smiling,
and we see a mixture of genders in the samples.
Further, while we see a greater diversity in the samples,
we also notice a slight drop in image quality (presumably because none of
the approaches has seen supervision with just `abstract' concepts).
See~\cref{subsec:more_celeba} for more qualitative examples on \CelebA.
}
On the top right, we show some examples of visual imagination,
where we ask the models to generate images from the concept
``bald female'', which does not occur in the training set.\footnote{
  There are 9 examples in the training set
  with the attributes (male=0, bald=1),
  but these turn out to all be labeling errors,
  as we shown in \cref{sec:celebAdetails}.
} %
(We omit the results from \bivcca, which are uniformly poor.)
We see that both \telbo and \jmvae
can sometimes do a fairly
reasonable job
(although these are admittedly cherry picked results).
Finally, the bottom right illustrates an interesting bias
in the dataset:
if we ask the model to generate images
where we do not specify the value of the
eyeglasses attribute, nearly all of the samples
fail to included glasses,
since the prior probability of this attribute is
rare (about 6\%).

\section{Concept Naming with Imagination Models}

In this section, we
demonstrate initial results which show that our imagination
models can be used for concept naming, where the task is
to assign a label to a set of images illustrating
the concept depicted by the images. A similar problem has been
studied in previous work such as~\citet{Tenenbaum99} and~\citet{Jia2013}.
\citet{Tenenbaum99} studies a set naming problem with integers (instead of images),
and show that
construct a likelihood function given a hypothesis set that can capture notions of
the minimal/smallest hypothesis that explains the observed samples in the set.
\citet{Jia2013} extend this approach to concept-naming on images, incorporating
perceptual uncertainty (in recognizing the contents of an image) using a confusion
matrix weighted likelihood term. While this approach first extracts labels for each
image and then performs concept naming, here we test how well our generative model
itself is able to generalize to concept naming without ever performing explicit
classification on the images.

In more detail, the problem setup
in concept naming is as follows: we are given as input a set
$\mathcal{X}$ of images, each of which corresponds to a concept
in the compositional abstraction hierarchy~\cref{fig:teaser}.
The task is to assign a label $\vy \in \mathcal{Y}$ to the set of
images. One of the key challenges in concept learning is to understand
``how far'' to generalize
in the concept hierarchy given a limited number of positive examples~\citep{Tenenbaum99}. That is, given a small set of images with 7 in the
top-left corner and bottom-right corner, one must infer that the
concept is ``7'' as opposed to ``7, top-left''. 
In other words, we wish to find the
least common ancestor (in the concept hierarchy) corresponding
to all the images in the set, given any number of images in the
set, so that we can be consistent with the set.
  We consider two heuristic solutions to this problem:
  
\begin{enumerate}
\item \textbf{Concept-NB}:
  In this approach we compute $\arg \max_{\vy} p(\vy| \mathcal{X})$,
  where $p(\vy| \mathcal{X})$ is computed using the naive bayes assumption:
	\begin{equation*}
	  p(\vy| \mathcal{X}) \propto  p(\vy) \Pi_{\vx_n \in \mathcal{X}} p(\vx_n| \vy)
          = p(\vy)  \Pi_{\vx_n \in \mathcal{X}} \int d\vz_n p(\vx_n| \vz_n) q(\vz_n|\vy)
	\end{equation*}
	where $p(y)$ is chosen to be uniform across
	all concepts, and the integrals are approximated using Monte Carlo.
%        $p(\vx| \vy)$ can be computed using a monte carlo%
%	approximation for $\int d\vz p(\vx| \vz) q(\vz| \vy)$ and . We
%        can then iterate over a feasible set of 
%	values for $\vy$ and assign a concept by maximizing $p(\vy| \mathcal{X})$.
  
	\item \textbf{Concept-Latent}: In this approach, instead
	  of working in the observed space,
          we work in the latent space. That is, we pick
          $\arg \min_{\vy} \KL(q(\vz| \mathcal{X})| q(\vz| \vy))$,
          where $q(\vz| \mathcal{X})$ is approximated using
          $\sum_{\vx \in \mathcal{X}} q(\vz| \vx)$, which is a mixture
          of gaussians. The 
	KL divergence can be computed analytically by considering the
	first two moments of the gaussian mixture\footnote{Given a Gaussian mixture of the form $g(\vx) = \sum_i \pi_i f(\vx; \mu_i, \sigma_i)$, where $f$ is the pdf for the Gaussian distribution, the first order moment, that is,
		the mean of $g(\vx)$ is given by: $\sum_i \pi_i \mu_i$. The variance is given by:
		$\sum_i \pi_i \sigma_i^2 + \sum_i \pi_i \mu_i^2 - (\sum_i \pi_i \mu_i)^2$.}.
\end{enumerate}

\subsection{Experimental Setup}
\ramacr{We use the \MNISTA dataset for the concept naming studies. We consider the fully specified attribute labels in the \MNISTA hierarchy, and consider differrent patterns of missingness (corresponding to different nodes in the abstraction hirearchy) by dropping attributes. Specifically, we ignore the case where no attribute is specified, and consider a uniform distribution over
the rest of the ($2^4-1=15$) patterns of missingness. Now, for
each fully specified attribute pattern in the iid split of \MNISTA,
we sample four missingness patterns and repeat across all
fully specified attributes to form a bank of 960 candidate names
that a model must choose. We randomly select three subsets of 100
candidate names (and the corresponding images) to form the query
set for concept naming, namely tuples of $(\vy, \mathcal{X})$.
Specifically, given all the images in the eval set for a concept $\vy$, we form $\mathcal{X}$ using a randomly sampled subset of
5 images. We report the accuracy metric, measuring how often the selected
concept for a set $\mathcal{X}$ matches the ground truth concept, across
three different splits of 100 datapoints.}

\subsection{Results}
\ramacr{We evaluate the best versions of \telbo, \jmvae, and \bivcca on the iid split of \MNISTA
for concept naming (\cref{table:concept}).
In general, we find that Concept-NB approaches perform significantly worse than Concept-Latent
approaches. For example, the best Concept-NB approach (using \telbo/\bivcca objective) gets to an accuracy
of around $18\%$, while Concept-Latent using \jmvae gets to $54.66 \pm 4.92\%$.
In general, these numbers are better than a random chance baseline which would get to $0.28\%$
(picking one of 348 effective options, after collating the 960 candidate names based on missingness patterns), while picking the most frequent (ground truth) fully-specified $\vy$ depicted across an image set gets to $6.33 \pm 1.88\%$.
\cref{fig:concept_naming} shows some qualitative examples from Concept-NB as well as
Concept-Latent models for concept / set classification. We observe that the Concept-Latent models are much more powerful than using Concept-NB in terms of naming
the concept based on few positive examples from the support set.
\begin{table} \footnotesize
	\setlength{\tabcolsep}{7.5pt}
	\tiny
	\begin{center}
		\begin{tabular}{@{} l  c  c @{}}
			%\hline
			\toprule
			Approach & Concept-Latent ($\%$) & Concept-NB ($\%$)\\
			\midrule
			\telbo & $35.66 \pm 2.05$ & $17.66 \pm 1.70$\\
			\jmvae & $54.66 \pm 4.92$ & $13.33 \pm 2.05$\\
			\bivcca & $28.00 \pm 4.54$ & $18.00 \pm 1.40$\\
			\midrule
			Random & $0.28 \pm 0.00$ & $0.28 \pm 0.00$\\
			Most Frequent & $6.33 \pm 1.88$ & $6.33 \pm 1.88$\\
			\bottomrule
			%Human & & & & & & & \\
			%\hline
		\end{tabular}
		\caption{Accuracy of Imagination models on Concept Naming. Higher is better.}
		\label{table:concept}
	\end{center}
\end{table}

\begin{figure}
	\centering
	\includegraphics[width=\textwidth]{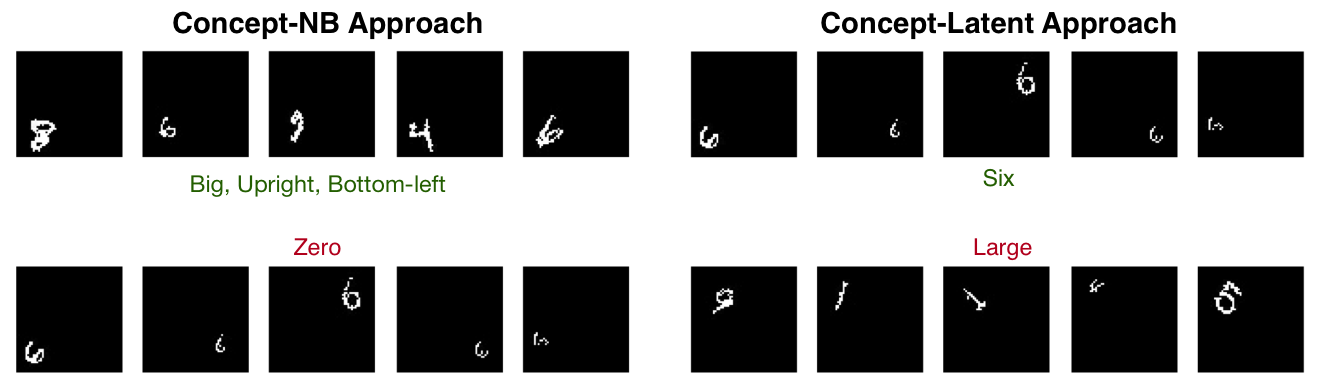}
	\caption{A qualitative illustration of some of the examples from concept naming models. Top-left: an example of a sample that is correctly named by a Concept-NB model. However, the Concept-NB model is not that
		strong and often gets simple concepts such as digits incorrect, making mistakes
		between 6 and 0, for example (bottom-left). This is likely because the only
		way in which the Concept-NB approach reasons about the set is not via a "meaningful" low dimensional latent variable but via a sampling distribution on a high dimensional space of images. The Concept-Latent model is able to do better
		on the same set of images, and classify the set as the concept ``6''. Finally,
		we show a failure case of the model where it incorrectly classifies the digits as being large (there is a small digit in the set), and ignores the fact that all of the digits are in the top-left.
	}
	\label{fig:concept_naming}
\end{figure}

\section{Conclusions and future work}
\label{sec:concl}

We have shown how to create generative
models which can ``imagine''
compositionally novel concrete and abstract  visual concepts.
In the future we would like to explore richer forms of description, beyond
attribute vectors, such as natural language text, as well as
compositional descriptions of scenes, 
 which will require dealing with a variable number of objects.

\subsubsection*{Acknowledgments}

We would like to thank
Hernan Moraldo for his help in writing
the JVAE library,
Alex Alemi for valuable insights on \telbo and JMVAE,
and Sergio Guadarrama and Harsh Satija for numerous discussions
around the project.
Finally we would like to thank Devi Parikh for advice
on the \CelebA experiments, and Stefan Lee and Yash Goyal
for feedback on an initial version of this draft.

\bibliographystyle{iclr2018_conference}
\bibliography{bib}

\newpage
\appendix
\section{Appendix}
\label{sec:appendix}
\label{sec:app}
\label{sec:suppl}

\subsection{Analysis of \JMVAE objective}
\label{sec:JMVAE}

The JMVAE objective of \citep{Suzuki2017} has the form
\[
J(\vx,\vy,\vtheta,\vphi)=
\elbo(\vx,\vy,\vtheta,\vphi)
-\alpha \left[ \KL(\qq(\vz|\vx,\vy), \qy(\vz|\vy))
+ \KL(\qq(\vz|\vx,\vy), \qx(\vz|\vx)) \right]
\]
Let us focus on the $\KL(\qq(\vz| \vx, \vy) | \qy(\vz| \vy))$ term.
Let $\calY$ be the set of unique labels (attribute vectors)
in the training set,
$\calX_i$ be the indices
of the images  associated with  label
$\vy_i$, and let $N_i=|\calX_i|$ be the size of that set.
Then we can write
\begin{align}
  \expectQ{\KL(\qq(\vz| \vx, \vy) | \qy(\vz| \vy))}{\pdata(\vx,\vy)}
  &= \frac{1}{|\calY|} \sum_{i \in \calY}
  \frac{1}{N_i} \sum_{n \in \vX_i}
  \KL(\qq(\vz| \vx_n, \vy_i), \qy(\vz| \vy_i))
\end{align}
As explained in \citep{elboSurgery},
we can rewrite this
by treating  the index $n \in \{1,\cdots, N_i\}$
as a random variable,
with prior $q(n|\vy_i)=1/N_i$.
Also, let us define the likelihood
$q(\vz|n,\vy_i) = \qq(\vz|\vx_n,\vy_i)$.
Using this notation, we can show that the above average KL becomes
\begin{align}
  \frac{1}{|\calY|} \sum_{i \in \calY}
  \left\{
  \KL(\qqavg(\vz|\vy_i) | \qy(\vz| \vy_i))
  + \log N_i
  -\expectQ{\entropy(q(n| \vz,\vy_i))}{\qy(\vz|\vy_i)}
  \right\}
  \label{eqn:elboJMVAE}
\end{align}
where
\[
\qqavg(\vz|\vy_i) =
\frac{1}{N_i}
\sum_{n \in \calX_i}
\qq(\vz|\vx_n,\vy_i)
\]
is the average of the posteriors for that concept,
and
$q(n|\vz,\vy_i)$ is
the posterior over the indices for all the possible
examples from the set $\calX_i$,
given that the latent code is $\vz$
and the description is $\vy_i$.
\eat{
We can compute this posterior as follows:
\begin{align}
  q(n|\vz,\vy_i)
  &= \frac{q(n|\vy_i) q(\vz|n,\vy_i)}{\sum_{n' \in \calX_i} q(n'|\vy_i) q(\vz|n',\vy_i)}
    = \frac{q(\vz|\vx_n,\vy_i)}{\sum_{n' \in \calX_i} q(\vz|\vx_{n'},\vy_i)}
\end{align}
where $q(\vz|n,\vy_i) = \qq(\vz|\vx_n,\vy_i)$
and $q(n|\vy_i) = 1/N_i$.
}

The $\KL(\qqavg(\vz|\vy_i) | \qy(\vz| \vy_i))$ term in
\cref{eqn:elboJMVAE}
tells us that JMVAE encourages the inference network for descriptions,
$\qy(\vz|\vy_i)$,
to be close to the average of the posteriors induced by
each of the images $\vx_n$ associated with $\vy_i$.
Since each $\qq(\vz|\vx_n,\vy_i)$ is close to a delta function
(since there is little posterior uncertainty when conditioning on an image),
we are essentially requiring that $\qq(\vz|\vy_i)$ cover the embeddings of each
of these images.

\subsection{Details on the \mnistaffine  dataset}
\label{sec:mnistDetails}

We created the \mnistaffine dataset as follows. Given an image in the
original MNIST dataset, we first sample a discrete scale label (big
or small), an orientation label (clockwise,  upright, and
anti-clockwise), and a location label (top-left, top-right,
bottom-left, bottom-right).

  Next, we converted this vector of discrete attributes
  into a vector of continuous transformation parameters, using the
  procedure described below:

\begin{itemize}
\item \textbf{Scale:} For big, we sample scale values from a
  Gaussian centered at 0.9 with a standard deviation of 0.1, while for
  small we sample from a Gaussian centered at 0.6 with a standard
  deviation of 0.1. In all cases, we reject and draw a sample again if
  we get values outside the range $[0.4, 1.0]$, to avoid artifacts
  from upsampling or problems with illegible (small) digits.
\item \textbf{Orientation:} For the clockwise label, we sample
  the amount of rotation to apply for a digit from a Gaussian centered
  at +45 degrees, with a standard deviation of 10 degrees. For
  anti-clockwise, we use a Gaussian at -45 degrees, with a standard
  deviation of 10 degrees. For upright, we set the rotation to be 0
  degrees always.
\item \textbf{Location:} For location, we place Gaussians at the
  centers of the four quadrants in the image, and then apply an offset
  of \texttt{image\_size/16} to shift the centers a bit towards the
  corresponding corners. We then use a standard deviation of
  \texttt{image\_size/16} and sample locations for centers of the
  digits. We reject and draw the sample again if we find that the
  location for the center would place the extremities of the digit
  outside of the canvas.
\end{itemize}

Finally, we generate the image as follows.
We first take an
  empty black canvas of size \texttt{64x64},  rotate the original
  \texttt{28x28} MNIST image, and then scale and translate the image and
  paste it on the canvas. (We use bicubic interpolation for scaling and
  resizing the images.)
  Finally, we use the method of~\citep{Murray2008} to binarize the
  images. See Figure~\ref{fig:mnistA} for example images generated in
  this way.

We repeat the above process of sampling labels, and applying
  corresponding transformations, to generate images 10 times for each
  image in the original MNIST dataset. Each trial samples labels from
  a uniform categorical distribution over the sample space for the
  corresponding attribute. Thus, we get a new \mnistaffine  dataset
  with 700,000 images from the original MNIST dataset of 70,000
  images. We split the images into a train, val and test set of 85\%,
  5\%, and 10\% of the data respectively to create the
  IID split. To create the compositional split, we split
  the \texttt{10x2x3x4=240} possible label combinations by the sample
  train/val/test split, giving us splits of the dataset with
  non-overlapping label combinations.

\begin{figure}[h]
  \centering
Co  \includegraphics[width=0.8\linewidth]{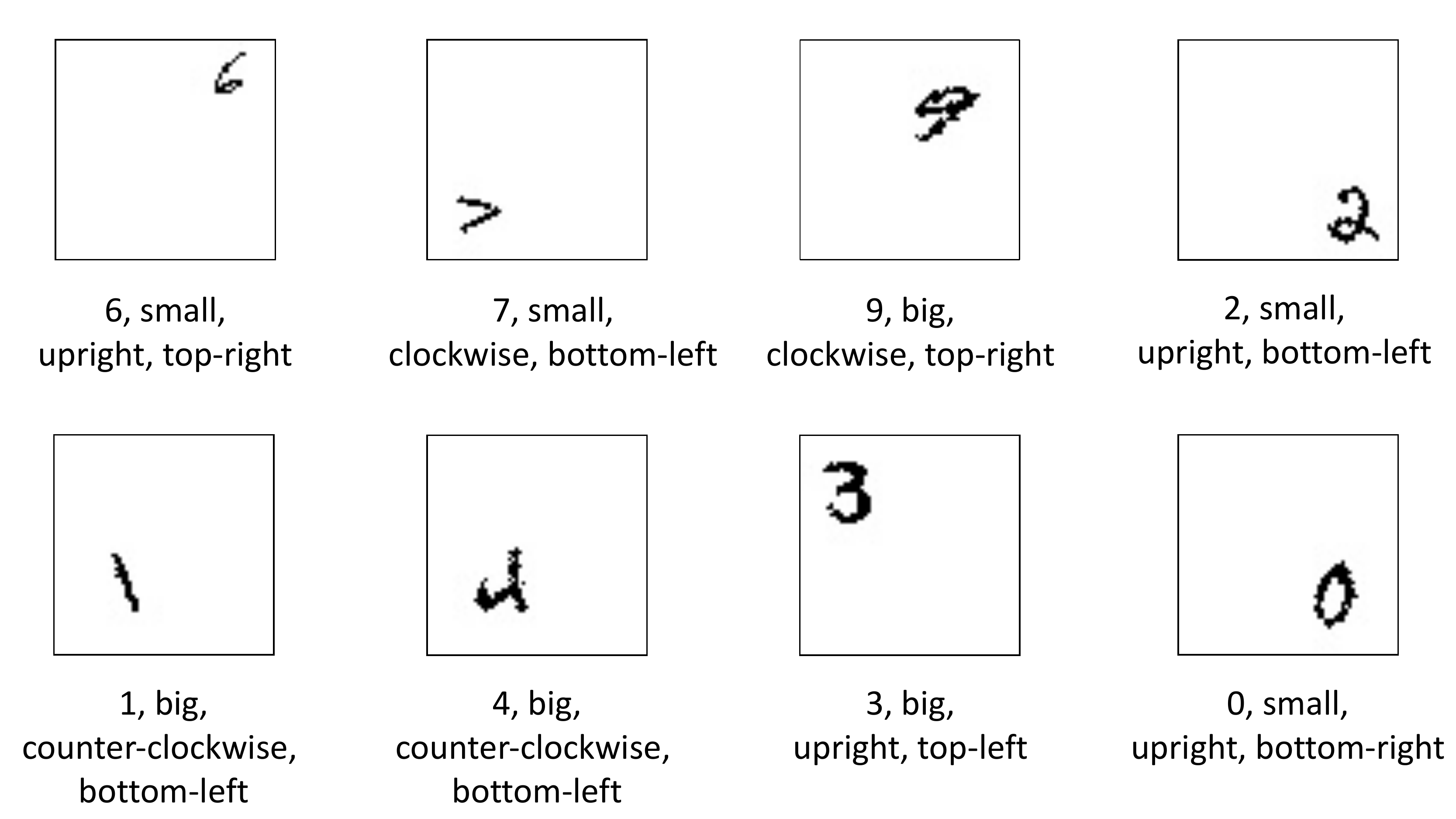}
  \caption{Example binary images from our \mnistaffine  dataset.}
  \label{fig:mnistA}
\end{figure}

%\subsection{$\beta$-VAE \vs likelihood scaling in joint models}
\subsection{$\beta$-VAE \vs Joint VAE}
\label{sec:mnist-latent}

\begin{figure}[ht]
	\centering \vspace{-3mm} \includegraphics[height=8pt,keepaspectratio]{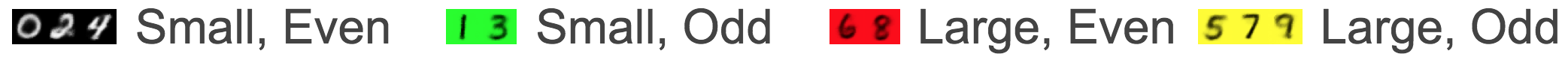}
	\begin{tabular}{ccc}
		\includegraphics[width=0.3\linewidth]{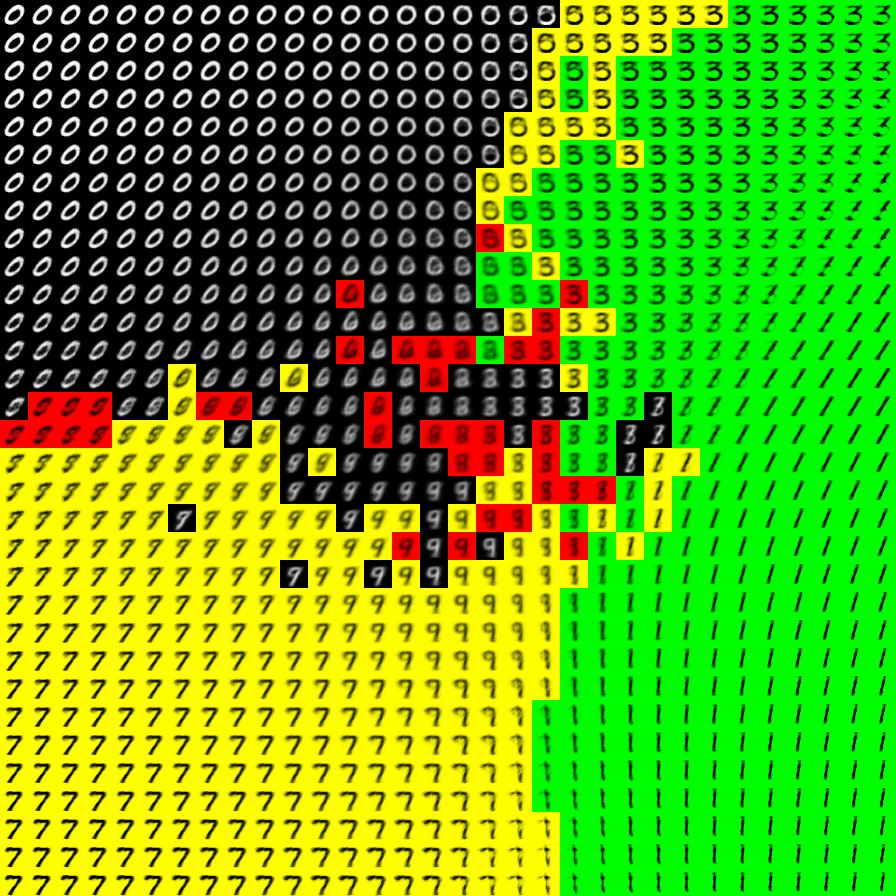} &
		\includegraphics[width=0.3\linewidth]{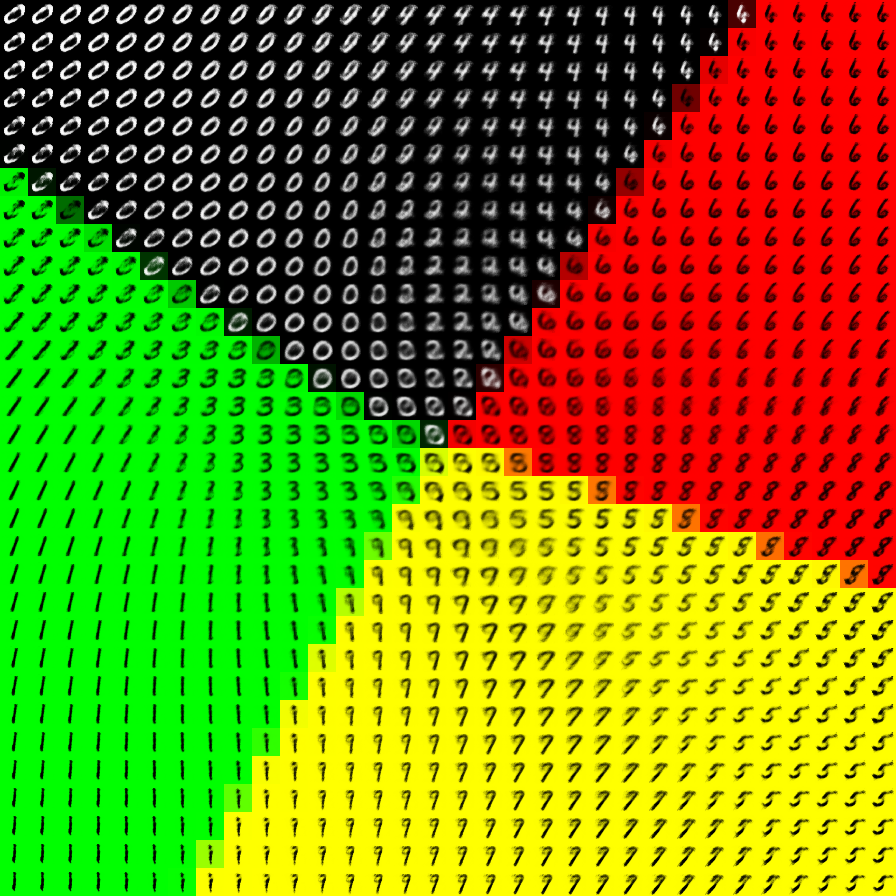} \\
		(a) & (b)
	\end{tabular}\vspace{-3mm}
	\caption{
		Visualization of the benefit of semantic annotations for learning a good latent space. Each small digit is a single sample generated
		from $p(x|z)$ from the corresponding point $z$ in latent space.
		(a) $\beta$-VAE fit to images without annotations.
		The color of a point $z$ is inferred from looking at the attributes of
		the training image that maps to this point of space using $q(z|x)$.
		Note that the red region  (corresponding to the concept of large and even digits)
		is almost non existent.
		% (b) Joint-VAE with $\lambda_y=1$.  Note that the red and green regions are disconnected.
		(b) Joint-VAE
                %with $\bimodaly=50$
                fit to images with annotations.
		The color of a point $z$ is inferred from $p(y|z)$.
	}
	\label{fig:geometry}
\end{figure}

$\beta$-VAE~\cite{Higgins2017} is an approach that aims to learn disentangled
latent spaces. It does this by modifying the ELBO objective,
so that it scales the $\KL(q(\vz|\vx),p(\vz))$
term by a factor $\beta > 1$.
This gives rise to disentangled spaces
since the prior $p(\vz)=\gauss(\vz|\vzero,\vI)$
is factorized
(see \citep{Achille2017} for details).
\eat{
tweaking the evidence lower bound to prioritize matching
the prior (which is assumed to be disentangled).
Our hypothesis/observation is
that one needs labels in order to truly disentangle and organize information
in latent spaces for high level concepts.
}
However, to learn latent spaces that correspond
to  high level concepts, this is not sufficient: we need to use labeled data
as well.

To illustrate this,  we set up
an experiment where we learn a 2d latent space for
standard MNIST digit images,
but where we replace the label
with two binary attributes:
parity (odd \vs even) and magnitude
(value $<5$ or $>=5$). We call this dataset MNIST-2bit.

In Figure~\ref{fig:geometry}(a), we show the results of fitting a 2d
$\beta$-VAE model \citep{Higgins2017} to the images in MNIST-2bit,
{\em ignoring the attributes}.  We perform a hyperparameter sweep over $\beta$, and
pick the one that gives the best looking latent space (this
corresponds to a value of $\beta=10)$.  At each point $z$ in the latent 2d
space, we show a single image sampled from  $p(x|z)$.
To derive the colors for each point in latent space, we proceed as
follows:
we embed each training image $x$ (with label $y(x)$) into latent
space, by computing
$\hat{z}(x) = E_{q(z|x)}[z]$. We then associate label
$y(x)$ with this point in space.
To derive the label for an arbitrary point $z$, we lookup the closest
embedded training image  (using  $\ell_2$ distance in $z$
space), and use its corresponding label.
We see that the latent space is useful for autoencoding
(since the generated images look good),
but it  does not capture the relevant semantic properties of
parity and magnitude.
In fact,
we argue that there is  no way of forcing the model to
learn a latent space that captures such high level conceptual
properties from images alone.

In Figure~\ref{fig:geometry}(b), we show the results of fitting a
joint VAE model to MNIST-2bit,  by optimizing $\elbo(x,y)$
on images and attributes
(\ie, we do not include the uni-modality $\elbo(x)$ and $\elbo(y)$ terms in this experiment.) 
Now the color codes are derived from
$p(y|z)$ rather than using nearest neighbor retrieval.  We see that
the latent space autoencodes well, and also captures the 4 relevant
types of concepts.
In particular, the regions are all convex and linearly seperable,
which facilitates the learning of a good imagination function
$q(z| y)$, interpolation, retrieval, and other latent-space
tasks.

A skeptic might complain that we have created an arbitrary
partitioning of the data, that is unrelated to the appearance of the
objects, and that learning  such concepts is therefore ``unnatural''.
But consider an agent interacting with an environment by touching
digits on a screen. Suppose
 the amount of reward they get depends on whether the digit
that they touch is small or big, or odd or even.
In such an environment, it would be very useful for the agent
to structure its internal representation to capture the concepts
of magnitude and parity, rather than in terms of low level visual
similarity.
(In fact,  \citep{Scarf2011} showed that pigeons can learn simple numerical
concepts, such as magnitude, by rewarding them for doing exactly this!)
Language can be considered as the realization of such concepts,
which enables agents to share useful information about their common
environments more easily.

\subsection{Details of the neural network architectures}
\label{sec:arch}

As explained in the main paper, we fit the joint graphical model
  $p(x, y,z) = p(z) p(x|z) p(y|z)$ with inference networks $q(z|
  x, y)$, $q(z|x)$, and $q(z|y)$. Thus, our overall model is made up
  of three encoders (denoted with $q$) and two decoders (denoted with
  $p$). Across all models we use the exponential linear unit
  (ELU)
  %~\citep{Clevert2015FastAA},
  which is a leaky non-linearity often used to train VAEs.
  %~\citep{Kingma2016iaf}.
  We explain the
  architectures in more detail below.

\textbf{\MNISTA model architecture}

\begin{itemize}

\item Image decoder, $p(x| z)$: Our architecture for the image decoder
  exactly follows the standard DCGAN architecture
  from~\citep{Radford2016}, where the input to the model is the latent
  state of the VAE.

\item Label decoder, $p(y| z)$: Our label decoder assumes a factorized
  output space $p(y|z) = \prod_{k \in \attributes} p(y_k|z)$, where $y_k$ is each
  individual attribute. We parameterize each $p(y_k| z)$ with a
  two-layer MLP with 128 hidden units each. We apply a small amount
  of $\ell_2$   regularization to the weight matrices.
  %on the first layer of the MLP, which consumes as
  %input the samples $z$ from inference networks.

\item Image and Label encoder, $q(z| x, y)$: Our architecture
(Figure~\ref{fig:q_z_xy}) for the
  image-label encoder first separately processes the images and the
  labels, and then concatenates them downstream in the network and
  then passes the concatenated features through a multi-layered
  perceptron.
  More specifically, we have convolutional layers which
  process image into \texttt{\(32, 64, 128, 16\)} feature maps with
  strides \texttt{\(1, 2, 2, 2\)} in the corresponding layers. We use
  batch normalization in the convolutional layers before applying the
  ELU non-linearity. On the label encoder side, we first encode the
  each attribute label into a 32d continuous vector and then pass each
  individual attribute vector through a 2-layered MLP with 512 hidden
  dimensions each.
  For example, for \mnistaffine we have 4 attributes, which
  gives us 4 vectors of 512d. We then concatenate these vectors and
  pass it through a two layer MLP. Finally we concatenate this
  label feature with the image feature after the convolutional layers
  (after flattening the conv-features) and then pass the result
  through a 2 layer MLP to predict the mean ($\mu$) and standard
  deviation ($\sigma$) for the latent space gaussian. Following
  standard practice, we predict $\log \sigma$ for the standard
  deviation in order to get values which are positive.

  \begin{figure}[ht]
  	\centering

  	\includegraphics[width=\columnwidth]{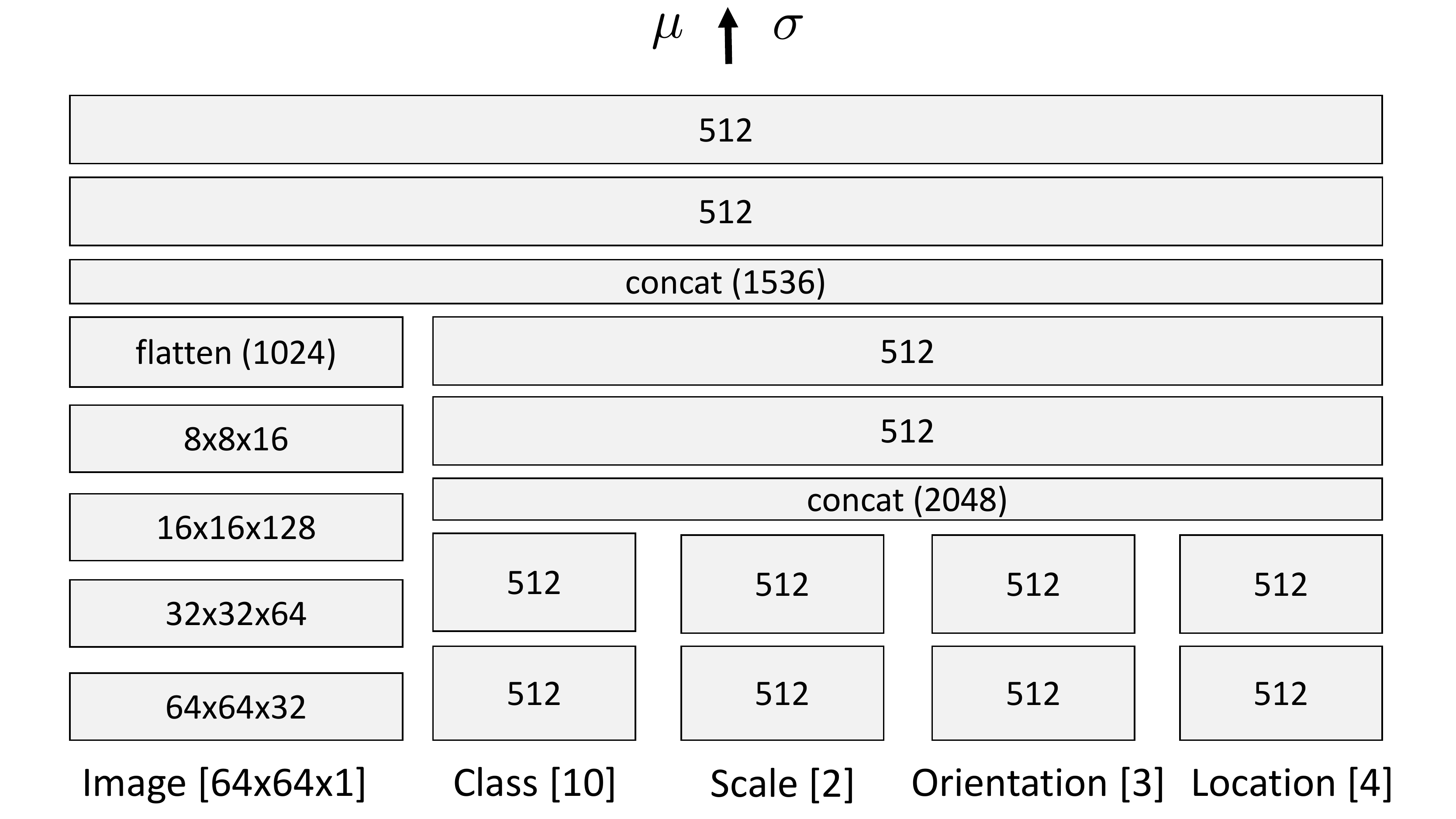}
  	\caption{Architecture for the $q(z| x, y)$ network in our JVAE models for \mnistaffine.
  		Images are \texttt{(64x64x1)}, class has 10 possible values,
  		scale has 2 possible values, orientation has 3 possible values,
		and location has 4 possible values.
  	}
    \label{fig:q_z_xy}
  \end{figure}

\item Image encoder, $q(z| x)$: The image encoder (Figure~\ref{fig:q_z_x}) uses the same
  architecture to process the image as the image feature extractor in
  $q(z|x, y)$ network described above. After the conv-features, we
  pass the result through a 3-layer MLP to get the latent state mean
  and standard deviation vectors following the procedure described
  above.

  \begin{figure}
\centering
\begin{subfigure}[b]{0.45\textwidth}
\centering
  	\includegraphics[width=\columnwidth]{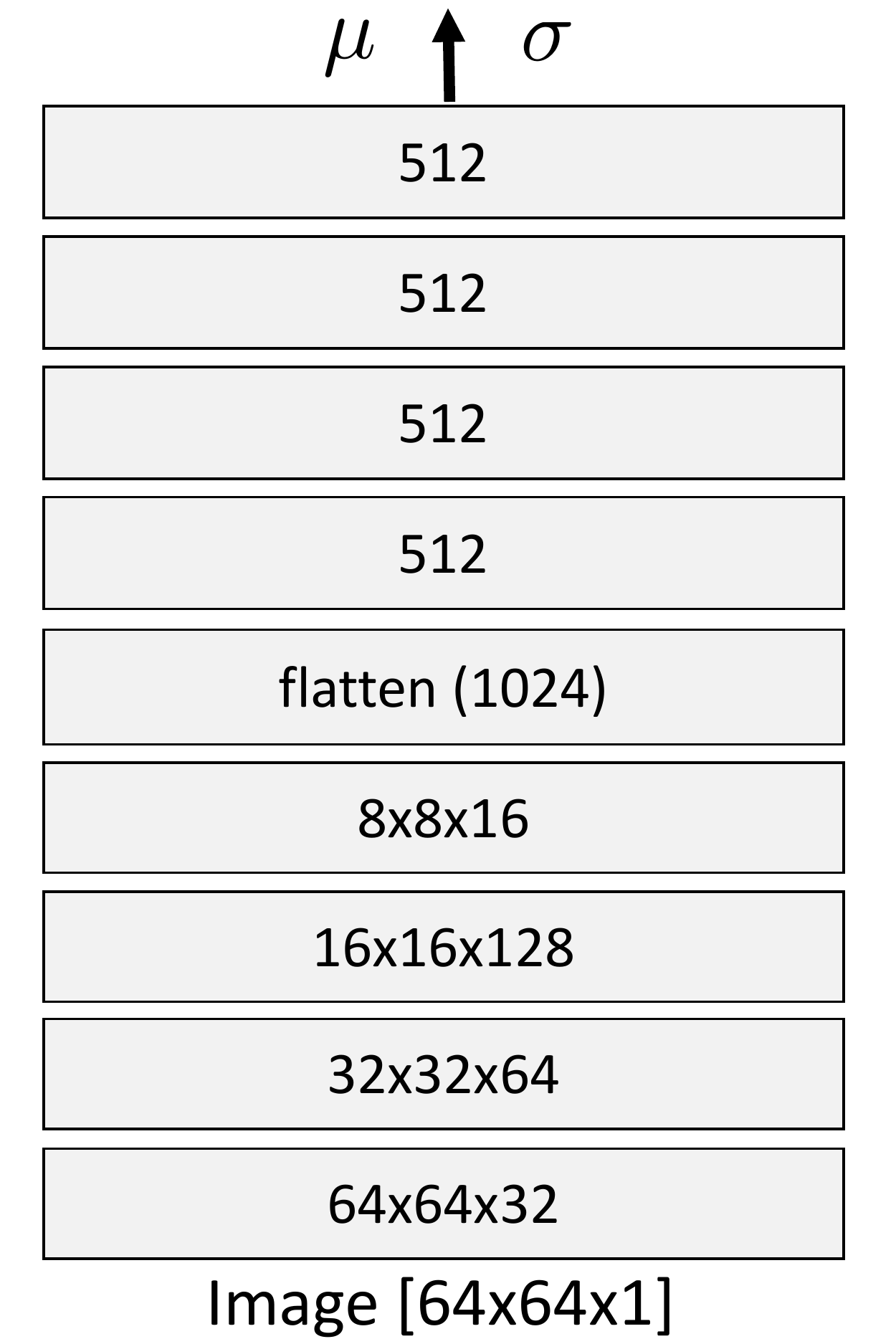}
  	\caption{Architecture for the $q(z| x)$ network.
  	}
    \label{fig:q_z_x}
  \end{subfigure}
~
\begin{subfigure}[b]{0.45\textwidth}
\centering
	\includegraphics[width=\columnwidth]{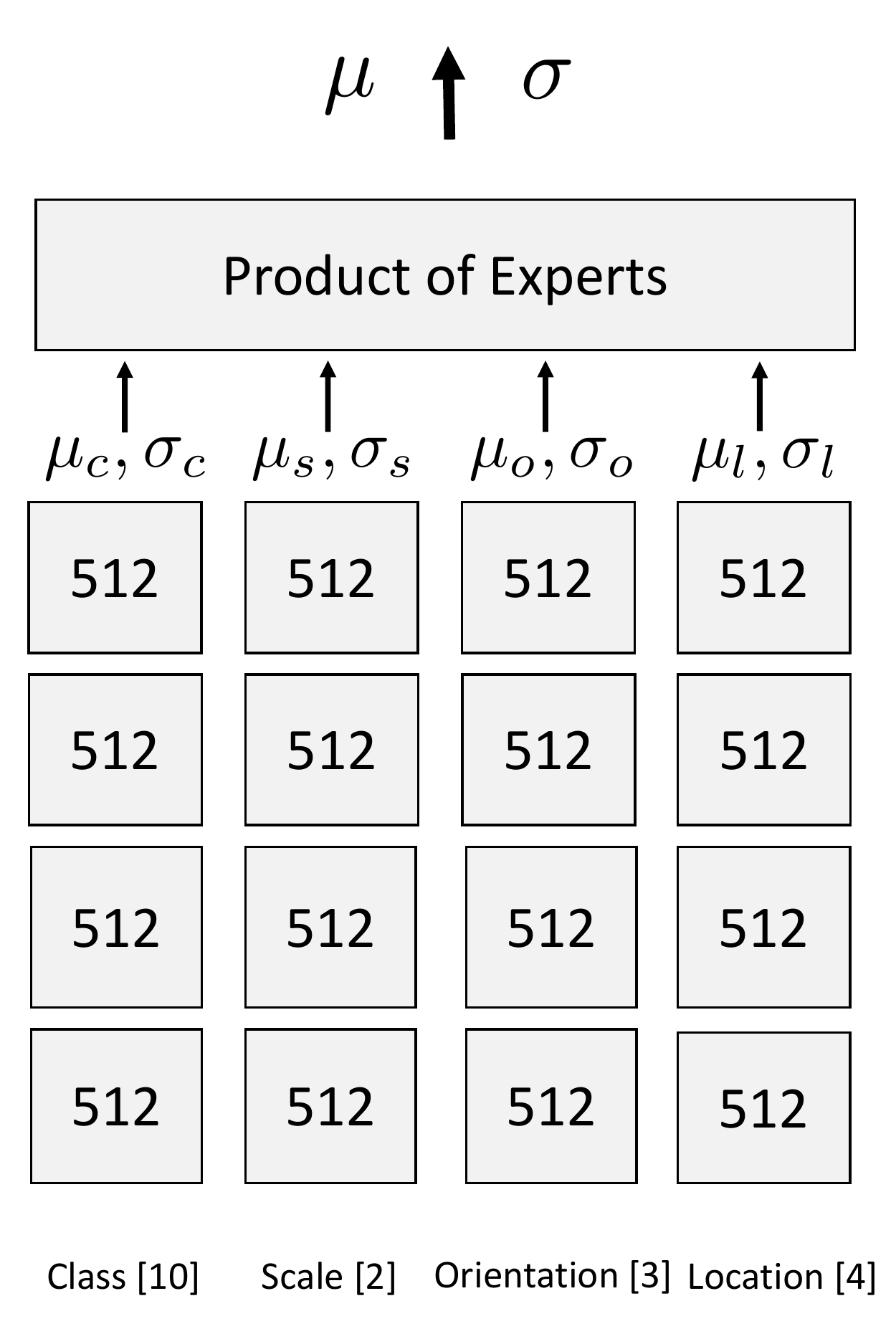}
	\caption{Architecture for the $q(z| y)$ network.
	}
	\label{fig:q_z_y_encoder}
\end{subfigure}
\caption{Archtectures for the single input inference networks
  for \mnistaffine.
  }
\label{fig:encoder}
\end{figure}

\item Label encoder, $q(z|y)$: The label encoder (Figure~\ref{fig:q_z_y_encoder}) part of the
architecture uses the same design choices to process the labels as
the label encoder part in the $q(z| x, y)$ network. After obtaining the initial, embeded
attributes, we pass the result through four distinct
4-layered MLPs with 512 hidden dimensions each and then
obtain the mean ($\mu$) and $\log \sigma$ values for each attribute in the
label set. The predicted $\mu$ and $\log \sigma$ values for each attribute are
fused together using the product of experts layer, which then outputs the
final parameters ($\mu$, $\log \sigma$) for the posterior.

\end{itemize}

\textbf{\mnistaffine Observation Classifier Model}
\rama{We next describe the architecuture of the observation classifier we
	use for evaluating the 3C's on the \mnistaffine dataset.}
\rama{The observation classifier is a convolutional neural network, with the first
convolutional layer with filters of size 5$\times$5, and 32 channels, followed
by a 2$\times$2 pooling layer applied with a stride of 2. This is followed by
another convolutional layer with 5$\times$5 filter size and 64 output channels.
This is followed by another 2$\times$2 pooling layer of stride 2. After this,
the network has four heads (corresponding to each attribute), each of which is an MLP with a single hidden layer (of
size 1024), with dropout applied to the activations. The final layer of the MLP
outputs the logits for classifying each attribute into the corresponding categorical
labels associated with it. We train this model from scratch on the \mnistaffine dataset
using stochastic gradient descent, batch size of 64 and a learning rate of $10^{-4}$.
}

\textbf{CelebA model architecture}
Our design choices for CelebA closely mirror the models we built for \MNISTA.
One primary difference is that we use a latent dimensionality of 18 in our
CelebA experiments which matches the number of attributes we model.
Meanwhile, the architectures of the image encoder, image decoder (\ie DCGAN),
are exactly identical to what is described above for \MNISTA execept that
encoders take as input a 3-channel RGB image, while decoders produce a 3-channel
output. We replace the Bernoulli likelihood with Quantized Normal likelihood
(which is basically gaussian likelihood with uniform noise).

In terms of the label encoder $q(z| y)$, we
follow Figure~\ref{fig:q_z_y_encoder} quite closely, except that we get as input
18 categorical (embedded) class labels as input, and we process the labels
through a single hidden layer before concatenation and two hidden layers post
concatenation (as opposed to two and four used in Figure~\ref{fig:q_z_y_encoder}).

Finally, the joint encoder $q(z| x, y)$, is again based heavily on Figure~\ref{fig:q_z_xy}
where we feed as input 18 labels as opposed to 4.

\subsection{Outputs of observation classifier on generated images}
\label{sec:obsClassifier}

\eat{
One of the advantages of using \MNISTA for studying visual abstraction is that
the sample quality on MNIST tends to be good enough for us to run classifiers
trained on real data on the images, and still expect to get reasonable
outputs. This allows us to calculate and measure our 3C's of visual imagination
in a systematic manner.
}

\cref{fig:observation_classifier_results} shows some images
sampled from our \telbo model trained on \MNISTA.
It also shows the attributes that are predicted by the attribute classifier.
 We see that the
		classifier often produces reasonable results that we as humans would also
		agree with. Thus, it acts as a reasonable proxy for humans classifying
		the labels for the generated images.

\begin{figure}[htbp]
	\centering
	\includegraphics[width=0.9\linewidth]{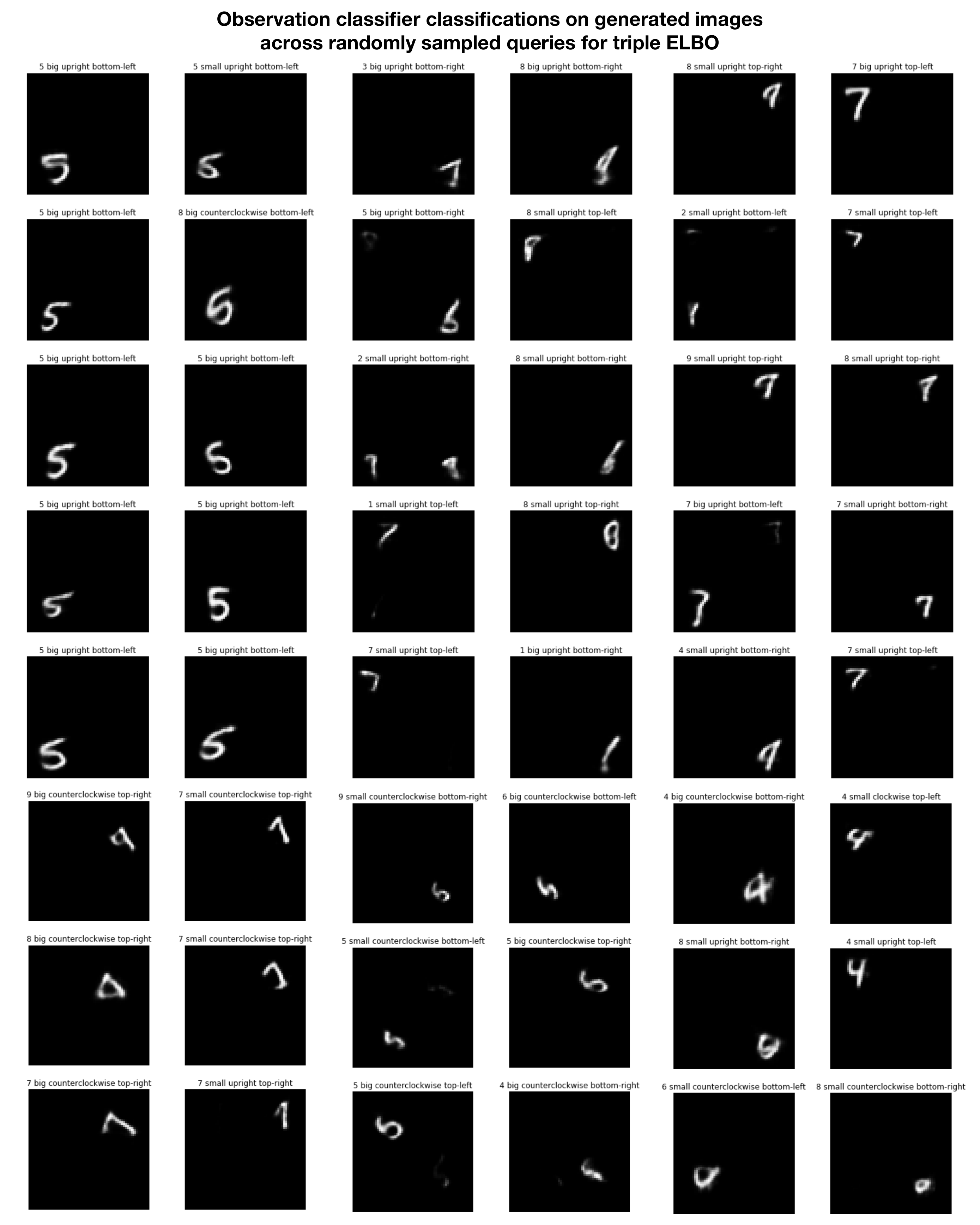}
	\caption{Randomly sampled images from the \telbo model
          when fed randomly sampled concepts from the \iid training set.
          We also show the outputs
	  of the observation classifier for the images.
          Note that we visualize mean
		images above (since they tend to be more human interpretable) but the
		classifier is fed samples from the model. Figure best viewed by zooming in.}
	\label{fig:observation_classifier_results}
\end{figure}

\subsection{Hyperparamter Choices for \telbo, \jmvae, \bivcca on \mnistaffine}
\rama{We discuss more hyperparameter choices for the different objectives and how
they impact performance on the \mnistaffine dataset. Across all the objectives
we set $\lambda_x$=1, and vary $\lambda_y$. In addition,
we also discuss how the private hyperparamter choices for each loss, $\gamma$ for
\telbo, $\alpha$ for \jmvae, as in~\cite{Wang16ccacca}) and $\mu$ for \bivcca affect performance. We use the JS-overall metric for 
picking hyperparameters, as explained in the main paper.
}

\begin{enumerate}
	\item Effect of $\lambda_y$: \rama{We search for $\lambda_y$ values in
		the set $\{1, 50, 100\}$ for all objectives. In general, we find the setting of $\lambda_y$
	in the $\elbo$ terms to be critical for good performance (especially on
	correctness). For example, at $\lambda_y$=1, we find that correctness
	numbers for the best performing \telbo model drop to 60.47 ($\pm$ 0.34) (from 82.08 ($\pm$ 0.56) at $\lambda_y$=50) on the validation set for \texttt{iid} queries.
	Similar trends can be observed
	for the \jmvae and \bivcca objectives as well (with $\lambda_y$=10 being the best
	setting for \bivcca, $\lambda_y$=50 for \jmvae). We have seen qualitative evidence which shows
	that the likelihood scaling for $\lambda_y$ affects how disentangled the
	latent space is along the specified attributes. When the latent space
	is not grouped or organized as per high-level attributes (see~\cref{fig:geometry} for example), the posterior distribution
	for a given concept is multimodal, which is hard for a gaussian inference
	network $q(\vz| \vy)$ to capture.
	This leads to poor correctness values.}

    \item Effect of $\gamma$: \rama{In addition to the $\lambda_y$ scaling term
    which is common across all objectives, \telbo has a $\gamma$ scaling factor
    which controls how we scale the $\log p(y|z)$ term in the
    $\elbo_{\gamma,1}(\vy,\vtheta_y,\vphi_{y})$ term. We sweep values of $\{1, 50, 100\}$ for this parameter. In general, we find that 
    the effect of this term is smaller on the performance than the $\lambda_y$
    term. Based on the setting of this parameter, we find that, for example,
    the correctness values for fully specified
    queries change from 82.08 ($\pm$0.56) at $\gamma$=50 to 80.27 ($\pm$0.38) at
    $\gamma$=1 on validation set for \texttt{iid} queries.}

    \item Effect of $\alpha$: \rama{We generally find that $\alpha$=1.0 works best
    for \jmvae across the different choices explored in~\cite{Wang16ccacca}, namely,
    $\{0.01, 0.1, 1.0\}$. For
    example, decreasing the value of $\alpha$ to 0.1 or 0.01 reduces correctness
    for fully sepcified queries
    from 85.63 ($\pm$0.29) to 77.58 ($\pm$0.23) at 0.1 and 74.57 ($\pm$0.44) at 
    0.01 respectively on the validation set for \texttt{iid} queries.}
    \item Effect of $\mu$: \rama{For \bivcca, we ran a search for $\mu$ over $\{0.3, 0.5, 0.7\}$, running each training experiment four times, and picked the best
    	hyperparameter choice across the runs.
    	We found that $\mu$=0.7 was the best value, however the performance
    	difference across different choices was not very large.
    	Intuitively, higher values of $\mu$ should lead to improved
    performance compared to lower values of $\mu$. This is because lower values
    of $\mu$ mean that we put more
    weight on the $\elbo$ term with a $q(\vz|\vx)$ inference network than the
    one with a $q(\vz|\vy)$ inference network, which results in sharper samples.
}
\end{enumerate}

\subsection{Compositional genralization on \MNISTA: Qualitative Results and Details}\label{subsec:comp_generalization_mnista}
We next show some examples of compositional generalization on \MNISTA on a validation set
of queries. For the compositinal experiments we reused the parameters of the best models
on the \iid splits for all the models, and trained the models for $\sim160K$ iterations.
All other design choices were the same. Figure~\ref{fig:compositional_generalization} shows some qualitative results.

\begin{figure}
	\centering
	\includegraphics[width=\textwidth]{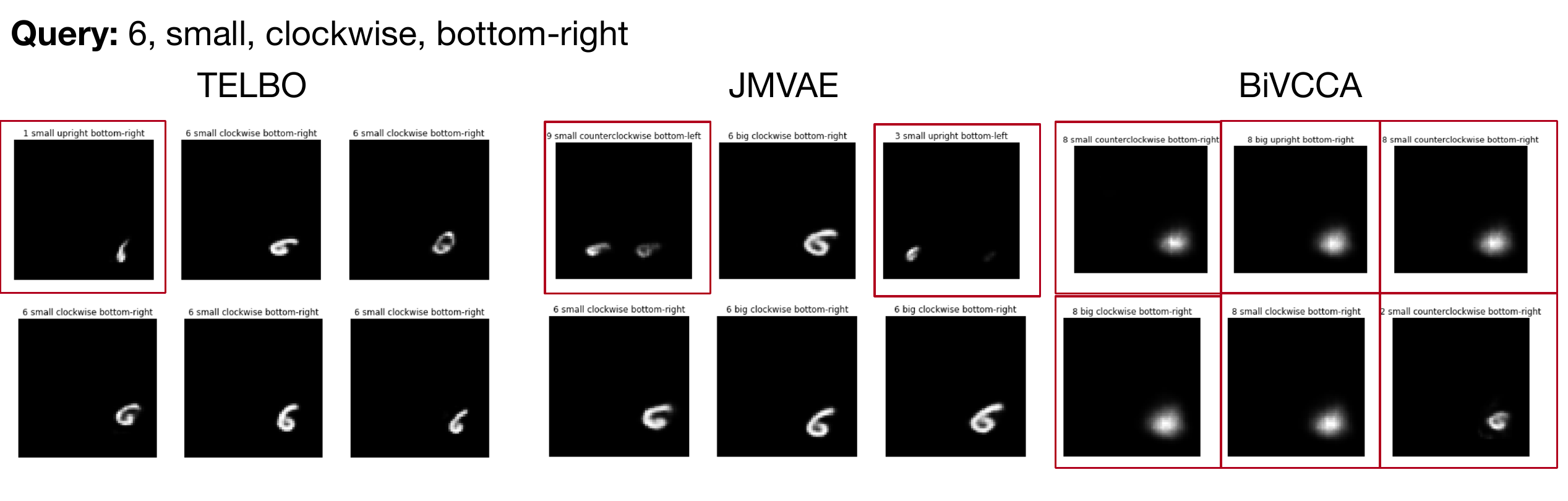}
	\caption{Compositional generalization on \MNISTA. Models are given the unseen
		compositional query shown at the top and each of the three columns shows
the mean of the image distribution generated by the models. Images marked with a red box
are those that the observation classifier detected as being incorrect. We also show
the classification result from the observation classifier on top of each image.
 We see that \telbo and \jmvae both do really well, while \bivcca is substantially poorer.
}
\label{fig:compositional_generalization}
\end{figure}

\subsection{Details on \CelebA}
\label{sec:celebAdetails}

\CelebA consists of 202,599 face colored images and 40
attribute binary vectors.
We use the version of this dataset
that was used in \citep{Perarnau2016};
this uses a subset of 18 visually distinctive attributes,
and preprocesses each image
so they are aligned, cropped, and scaled down to 64 x 64.
We use the official train and
test partitions, 182K for training and 20K for testing.
Note that this is an \iid split,
so the attribute vectors in the test set
all occur in the training set,
even though the images and people are unique.
In total, the original dataset with 40 attributes specified
a set of 96486 unique visual concepts, while our dataset of
18 attributes spans 3690 different visual concepts.

In \cref{sec:celebAresults},
we claim that our generations of ``Bald'' and ``Female''
images are from a compositionally novel concept.
Our claim comes with
a minor caveat/clarification:
the concept
\texttt{bald}$=$1 and \texttt{male}$=$0 does occur in 9 training examples,
but they  are all
incorrect labelings,
as shown in \cref{fig:baldFemales}!
Further, we see that the images generated from our model
(shown in \cref{fig:celeba}) are qualitatively very different from any
of the images here,
showing that the model has not memorized these examples.

\begin{figure}
	\centering
	\includegraphics[width=\textwidth]{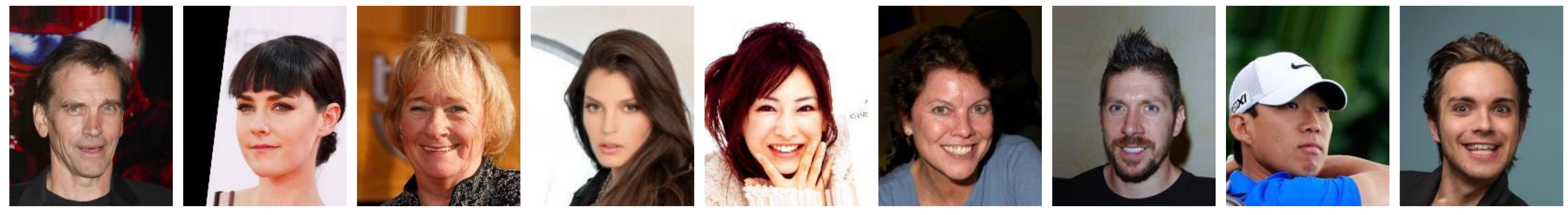}
\caption{Set of all 9 images labelled as \texttt{bald}$=$1 and \texttt{male}$=$0
	in the \CelebA dataset. We can see that in all the cases the labels are
	inaccurate for the image, probably due to annotator error.
}
\label{fig:baldFemales}
\end{figure}

\begin{figure}[tbp]
	\centering
	\includegraphics[width=0.6\textwidth]{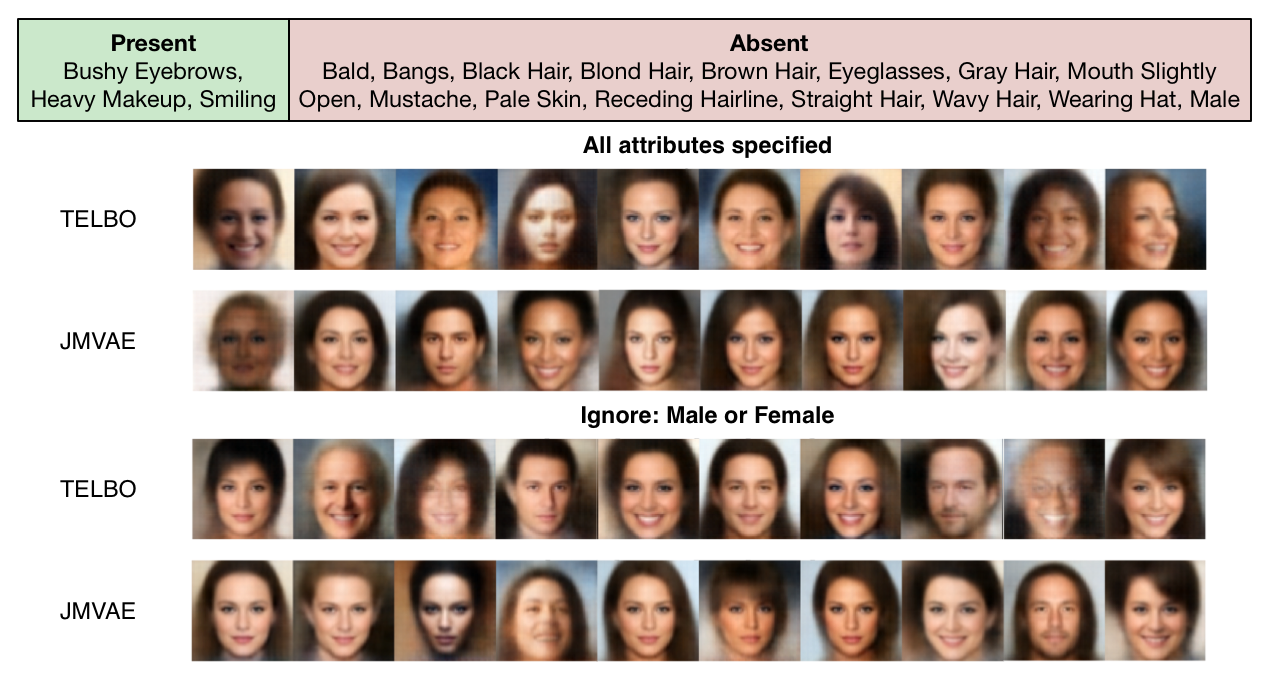}	
	\caption{\rama{\textbf{\telbo creates more diverse images than \jmvae.} At the top we
			show the set of attributes which are present and absent in the input
			query. Below, we show the results of generation with all the attributes
			specified, drawing 10 samples each.
			We see that both \telbo and \jmvae create accurate images
			satisfying the constraints. Note that the concept ``male'' is set to ``absent'' in the query,
			which in \CELEBA means that ``female'' is present. Next, we unspecify
			whether the image should contain a male or a female. We see that in
			this setting, \telbo has a better mixing of male and female images
			(fourth, sixth, eighth and ninth images in the third row are male),
			than \jmvae which just produces a single male image (the ninth image
			in the fourth row).}
	}
	\label{fig:celeba_qualitative_diversity}
\end{figure}

\subsection{More results on \CELEBA}\label{subsec:more_celeba}
\rama{Finally, we show further qualitative examples of performance
	on the \CELEBA dataset. We focus on the \telbo and \jmvae objectives
	here, since \bivcca generally produces poor samples (see~\cref{fig:celeba}).~\cref{fig:celeba_qualitative_diversity} (middle) shows some example generations for the concept specified by the attributes (top). We see that both \telbo and \jmvae produce correct images
	when provided the full attribute queries (first two rows). However, when we
	stop specifying attribute
	``male'' or ``not male'' (female), we see that \telbo provides
	more diverse samples, spanning both male and female (compared to \jmvae).
	This ties into the explanation in~\cref{sec:JMVAE}, where we show
	how one can interpret JMVAE as optimizing for the $\KL(\qqavg(\vz|\vy_i) | \qy(\vz| \vy_i))$
	to fit the unimodal inference network $\qy(\vz|\vy_i)$. Since \jmvae
	only reasons about the ``aggregate'' posterior as opposed to the prior
	(which \telbo reasons about), it has the tendency to generate less diverse samples when shown unseen concepts.
}

\end{document}